\documentclass[sigconf, nonacm]{acmart}

\newcommand\vldbdoi{XX.XX/XXX.XX}
\newcommand\vldbpages{XXX-XXX}
\newcommand\vldbvolume{14}
\newcommand\vldbissue{1}
\newcommand\vldbyear{2020}
\newcommand\vldbauthors{\authors}
\newcommand\vldbtitle{\shorttitle} 
\newcommand\vldbavailabilityurl{URL_TO_YOUR_ARTIFACTS}
\newcommand\vldbpagestyle{plain}

\usepackage[ruled]{algorithm2e}
\usepackage{subfigure}
\usepackage{multicol}
\usepackage{multirow}
\usepackage{colortbl}
\definecolor{mycolor}{HTML}{E0E7F9}

\newtheorem{problem}{Problem}
\newtheorem{definition}{Definition}
\newtheorem{theorem}{Theorem}

\newcommand{\eg}{\emph{e.g.},\xspace}
\newcommand{\ie}{\emph{i.e.},\xspace}

\begin{document}
\title{Against Multifaceted Graph Heterogeneity via Asymmetric Federated Prompt Learning}

\author{Zhuoning Guo}
\email{zguo772@connect.hkust-gz.edu.cn}
\affiliation{%
  \institution{The Hong Kong University of Science and Technology (Guangzhou)}
  \country{}
}

\author{Ruiqian Han}
\email{rhan464@connect.hkust-gz.edu.cn}
\affiliation{%
  \institution{The Hong Kong University of Science and Technology (Guangzhou)}
  \country{}
}

\author{Hao Liu}
\email{liuh@ust.hk}
\affiliation{%
  \institution{The Hong Kong University of Science and Technology (Guangzhou)}
  \institution{The Hong Kong University of Science and Technology}
  \country{}
}

\begin{abstract}
Federated Graph Learning (FGL) aims to collaboratively and privately optimize graph models on divergent data for different tasks. A critical challenge in FGL is to enable effective yet efficient federated optimization against multifaceted graph heterogeneity to enhance mutual performance. However, existing FGL works primarily address graph data heterogeneity and perform incapable of graph task heterogeneity. To address the challenge, we propose a Federated Graph Prompt Learning (FedGPL) framework to efficiently enable prompt-based asymmetric graph knowledge transfer between multifaceted heterogeneous federated participants. Generally, we establish a split federated framework to preserve universal and domain-specific graph knowledge, respectively. Moreover, we develop two algorithms to eliminate task and data heterogeneity for advanced federated knowledge preservation. First, a Hierarchical Directed Transfer Aggregator (HiDTA) delivers cross-task beneficial knowledge that is hierarchically distilled according to the directional transferability. Second, a Virtual Prompt Graph (VPG) adaptively generates graph structures to enhance data utility by distinguishing dominant subgraphs and neutralizing redundant ones. We conduct theoretical analyses and extensive experiments to demonstrate the significant accuracy and efficiency effectiveness of FedGPL against multifaceted graph heterogeneity compared to state-of-the-art baselines on large-scale federated graph datasets.
\end{abstract}

\maketitle

\pagestyle{\vldbpagestyle}
\begingroup\small\noindent\raggedright\textbf{PVLDB Reference Format:}\\
\vldbauthors. \vldbtitle. PVLDB, \vldbvolume(\vldbissue): \vldbpages, \vldbyear.\\
\href{https://doi.org/\vldbdoi}{doi:\vldbdoi}
\endgroup
\begingroup
\renewcommand\thefootnote{}\footnote{\noindent
This work is licensed under the Creative Commons BY-NC-ND 4.0 International License. Visit \url{https://creativecommons.org/licenses/by-nc-nd/4.0/} to view a copy of this license. For any use beyond those covered by this license, obtain permission by emailing \href{mailto:info@vldb.org}{info@vldb.org}. Copyright is held by the owner/author(s). Publication rights licensed to the VLDB Endowment. \\
\raggedright Proceedings of the VLDB Endowment, Vol. \vldbvolume, No. \vldbissue\ %
ISSN 2150-8097. \\
\href{https://doi.org/\vldbdoi}{doi:\vldbdoi} \\
}\addtocounter{footnote}{-1}\endgroup

\ifdefempty{\vldbavailabilityurl}{}{
\vspace{.3cm}
\begingroup\small\noindent\raggedright\textbf{PVLDB Artifact Availability:}\\
The source code, data, and/or other artifacts have been made available at \url{\vldbavailabilityurl}.
\endgroup
}

\section{Introduction}

Federated Graph Learning~(FGL) has become an attractive research direction for distributed Graph Neural Network~(GNN) optimization on isolated graph data without explicit information exposure~\cite{liu2024federated}. Specifically, in FGL, several participants maintain their private graphs collected from different domains, and they aim to collaboratively train GNNs that can provide beneficial knowledge of raw graph data for particular downstream tasks, respectively. 

Recent FGL studies have achieved significant progress mainly in learning more effective graph representation to enhance downstream task prediction~\cite{guo2024hifgl}. For example, FedSage+~\cite{zhang2021subgraph} is proposed based on a standard FGL pipeline where subgraphs from different silos are independently distributed, and improve the node classification via missing edge generation.
Despite these advancements, the difficulty of graph data heterogeneity remains paramount. Graphs inherently differ in terms of node and edge types, structural configurations, and other characteristics. Traditional graph learning methodologies often operate under the assumption of homogeneity, which can lead to biased or suboptimal models when applied to heterogeneous graph domains. Addressing graph data heterogeneity is crucial for FGL, as it enables the model to capture diverse intrinsic graph characteristics and enhances overall performance.
Therefore, FedStar~\cite{tan2023federated} investigates the data heterogeneity of graph features and shared knowledge, which attempts to achieve outperformance in a Non-Independent and Non-Identically Distributed~(Non-IID) issue setting. In addition, FedLIT~\cite{xie2023federated} detects link-type heterogeneity to decrease the harmful structure to allow more beneficial message passing. Generally, these works designed effective methods to tackle the graph data heterogeneity in terms of features and structures, where they demand participants to train their models for a consistent downstream task.

However, while attempts have been made to address data heterogeneity, the critical issue of task heterogeneity remains unexplored.
FGL participants originate from various sectors, each with its own set of tasks and data characteristics. For instance, healthcare providers may focus on patient data analysis, while financial institutions might prioritize fraud detection. Addressing task heterogeneity allows FGL to adapt to these varied needs, enhancing its applicability and relevance across different applications.
Existing federated algorithms, which typically assume a uniform model structure across clients, cannot accommodate the need for diverse architectures in graph learning~\cite{sun2023graph}. This limitation complicates parameters and knowledge sharing between clients working on different tasks, potentially causing inferior effectiveness and efficiency in collaborative optimization.
To effectively optimize models against multifaceted heterogeneity, we necessitate a system not only to preserve common knowledge adaptable to various tasks but also to generalize across divergent graph data distributions.
Therefore, the core challenge of this work is to simultaneously overcome the multifaceted heterogeneity for more effective and efficient FGL.

To address the challenge, we propose a \textbf{Federated Graph Prompt Learning~(FedGPL)} framework to federally fine-tune graph models on heterogeneous tasks and data via an efficient prompt-based asymmetric knowledge transfer.
Overall, we establish a split framework to simultaneously achieve a universal graph representing and personalized graph prompting to preserve global and local knowledge, respectively.
Then, we design tailored algorithms to disentangle and address task and data heterogeneity.
On the server side, we develop a Hierarchical Directed Transfer Aggregator~(HiDTA) to extract and share asymmetrically beneficial knowledge among task-heterogeneous participants via personalized federated aggregation.
On the client side, we devise a lightweight prompting module, called Virtual Prompt Graph~(VPG), to adaptively generate augmented graph data by distilling more dominant information with minor data heterogeneity.
We provide theoretical analyses that demonstrate the effectiveness in reducing task and data heterogeneity, as well as the significant reduction of memory and communication costs.
Extensive experiments validate that FedGPL outperforms baseline methods across three levels of tasks against multifaceted graph heterogeneity on five datasets.
Notably, we evaluate FedGPL in a typical large-scale FGL system consisting of $1$ million node data. Our method achieves $5.3\times \sim 6.0\times$ GPU memory efficiency, $2.1\times \sim 3.7\times$ communication efficiency, and $1.3\times \sim 1.9\times$ training time efficiency. The efficiency superiority demonstrates its scalability for massive FGL participants and large-scale graph data.

Our main contributions can be concluded as:
(1)~To our knowledge, our work is the first to study both task and data heterogeneity in FGL, addressing gaps overlooked by previous research.
(2)~We propose a federated graph prompt learning framework to effectively enable federated optimization of personalized models among task- and data- heterogeneous participants.
(3)~We develop an aggregation algorithm to deliver task-specific knowledge based on a transferability-aware hierarchy to asymmetrically enhance model performance. Moreover, we devise a virtual graph prompt to jointly highlight dominant graph structures and alleviate extensive data heterogeneity.
(4)~We theoretically analyze the mitigation of multifaceted heterogeneity by FedGPL. Besides, we conduct extensive experiments to prove the accuracy and efficiency superiority of FedGPL against FGL with multifaceted heterogeneity on federated graph datasets with million nodes.

\section{Preliminaries}

\subsection{Notations and Problem Statement}

\subsubsection{Graph notations.}
Graph is a common data structure representing a set of objects and their relationships.
We denote a graph as $G=(V, E, X)$. Here $V=\{v_1, v_2, \cdots, v_{N_n}\}$ is the node set with $N_n$ nodes. $E=\{e_1, e_2, \cdots, e_{N_e}\}$ is the edge set with $N_e$ edges that $e_i=(v_{s_i}, v_{t_i})$ is an undirected edge from node $v_{s_i}$ to node $v_{t_i}$. $X\in \mathbb{R}^{N_n\times d_0}$ is the node feature matrix that $i$-th row vector $x_i\in \mathbb{R}^{d_0}$ is the feature of $v_i$, where $d_0$ is the node feature dimension.
For graph learning, researchers usually treat three levels~(\ie node-level, edge-level, and graph-level) of tasks in different pipelines~\cite{sun2023all}. We formulate three levels of tasks in Appendix~\ref{sec:appendix_graph_task}.

\subsubsection{Multifaceted heterogeneity.}
This paper investigates an FGL scenario consisting of multifaceted heterogeneity in terms of task and data heterogeneity, which is defined below.
\begin{definition}\label{def:task_het}
    \textbf{Task Heterogeneity.} In FGL, task heterogeneity arises when participants aim to optimize models for different downstream tasks, including node-, edge-, and graph- level tasks. Therefore, task heterogeneity will be reflected by the difference of model structures and parameters. Specifically, we measure the heterogeneity $\Delta_{T}^{a,b}(\theta_a,\theta_b) \in [0,1]$ between $a$-th and $b$-th tasks as
    \begin{equation}\label{equ:task_het}
        \Delta_{T}^{a,b}(\theta_a,\theta_b)=\left\{\begin{array}{cc}
            1, & t_a \neq t_b, \\
            \frac{1-\exp(-\overline{\Vert \theta_a - \theta_b\Vert_2})}{1+\exp(-\overline{\Vert \theta_a - \theta_b\Vert_2})}, & t_a = t_b,
        \end{array}\right.
    \end{equation}
    where $\theta_a$, $\theta_b$ denotes the model parameters for the $a$ -th and $b$ -th tasks, $\overline{\theta}$ means the average values of $\theta$.
\end{definition}
\begin{definition}\label{def:data_het}
    \textbf{Data Heterogeneity.} In FGL, data heterogeneity appears when participants keep divergent graph data generated from their domains. The data heterogeneity between graphs is reflected in the difference of their structures and features. Graph representations can incorporate graph structures and features in a high-dimensional vector.
    We utilize a popular graph readout function~\cite{xu2018powerful} to produce graph representations as $ H = \{h_1, \cdots, h_n\} = \mathcal{G}(V,E,X)$ and $h_{G} = \tilde{H} = \frac{1}{n}\sum^{i=1}_{n}h_i$, where $n$ is the number of nodes of $G$ and $\mathcal{G}$ is a trained GNN to represent nodes.
    Hence, we measure the data heterogeneity $\Delta_{D}^{i,j}(h_{G_i},h_{G_j}) \in [0,1)$ between graphs $G_i$ and $G_j$ according to the distance of their representations $h_{G_i}$ and $h_{G_j}$ as
    \begin{equation}\label{equ:data_het}
        \Delta_{D}^{i,j}(h_{G_i},h_{G_j}) = \frac{1-\exp(\overline{(h_{G_i} - h_{G_j})^2})}{1+\exp(\overline{(h_{G_i} - h_{G_j})^2})},
    \end{equation}
    where $\overline{(h_{G_i} - h_{G_j})^2}$ is the average value among $d$ dimensions of $(h_{G_i} - h_{G_j})^2$.
\end{definition}

\subsubsection{Problem formulation.}
\begin{problem}
    \textbf{Federated Graph Learning against Multifaceted Heterogeneity.} An FGL system consists of a server $S$ and $N$ clients $\{C_1,C_2,\cdots,C_N\}$ corresponding to each participant. Each client aims to learn a model $\mathcal{F}_i(\cdot)$ on its dataset $D_i$, where $t_i \in \{node, edge, graph\}$ and $1 \leq i \leq N$.
    Models and datasets may be heterogeneous among clients.
    We aim to optimize the parameters of all models to minimize the average loss as
    \begin{equation}
        \{\theta_1, \cdots, \theta_N\} = {\arg\min} \sum\limits^{i=1}_{N}\mathcal{L}_i(\mathcal{F}_i(D_i; \theta_i)),
    \end{equation}
    where $\mathcal{L}_i(\cdot)$ denotes the loss function of the prediction model $\mathcal{F}_i(\cdot))$ parameterized with $\theta_i$ on dataset $D_i$.
\end{problem}

\begin{figure*}
    \centering
    \includegraphics[width=\linewidth]{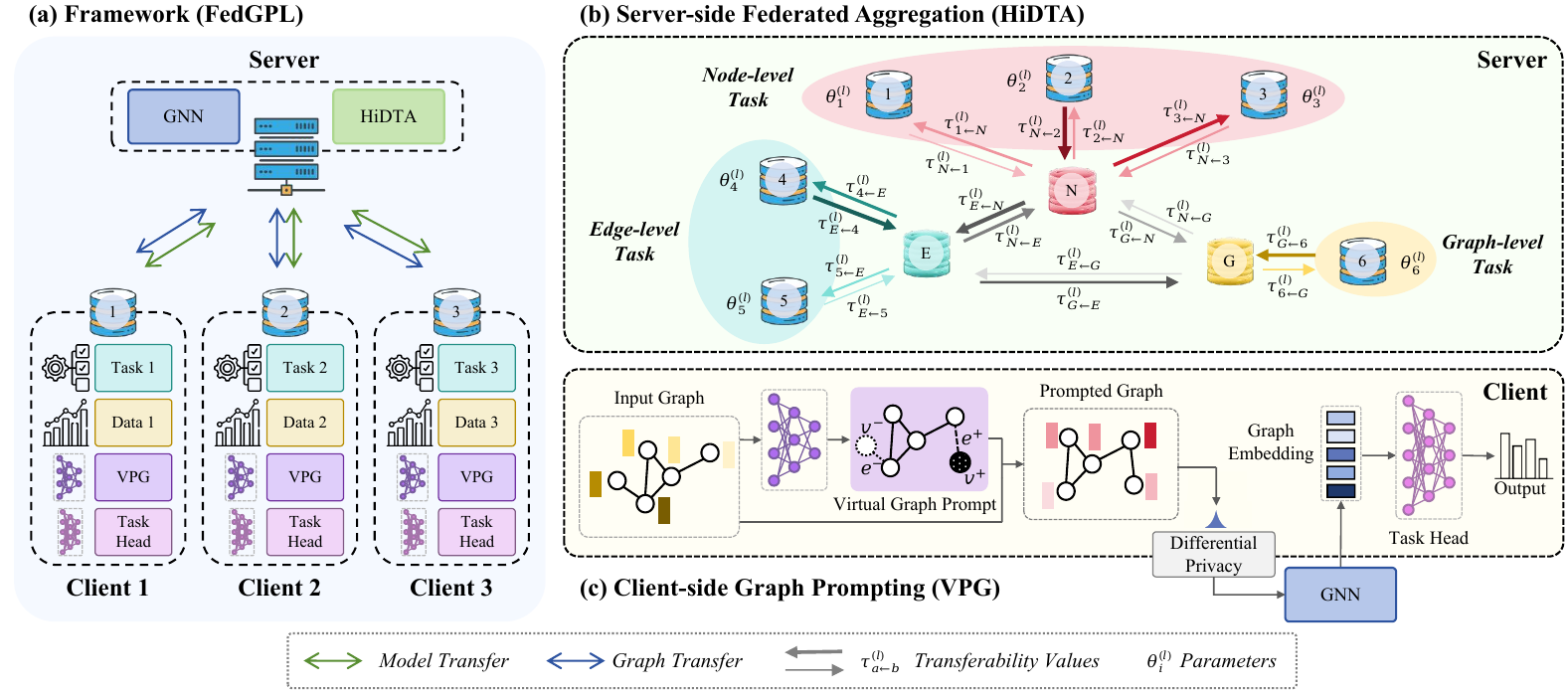}
    \caption{An overview of Federated Graph Prompt Learning~(FedGPL) framework.
    }
    \label{fig:framework}
\end{figure*}

\subsubsection{Threat model.}
We assume all participants are semi-honest, strictly following the Federated Learning~(FL) protocol, and only knowing about the knowledge they already keep or receive from others during interactions. We will clarify the objectives of participants, including clients and server: The client aims to obtain the structure and parameters of GNN from the server; the server aims to see the private data of the client.

\subsection{Graph Prompt Learning}\label{sec:gpl}
We formulates the pipeline of Graph Prompt Learning~(GPL), which can enable effective optimization for downstream tasks to inherit knowledge of Graph Neural Network~(GNN) and fine-tune on specific domains.

We first follow the existing work~\cite{sun2023all} to reformulate node-level and edge-level tasks as graph-level tasks to train three tasks in a consistent workflow illustrated in Appendix~\ref{sec:appendix_graph_ind}.
Then, GNN aims to produce representations by encoding input graphs.
Existing works pre-train GNNs on large-scale graph data for efficient fine-tuning and knowledge reusing~\cite{xia2022survey,rong2020self,zhu2023dual}. In this work, we leverage pre-trained GNNs in GPL for better performance.
A GNN $\mathcal{G}$ reads an input graph $G=(V, E, X)$ and outputs representations $H$ as $\mathcal{G}(G)=H \in \mathbb{R}^{N_n \times d}$, where $d$ is the dimension.

Specifically, we can insert a prompt into a graph from a source domain, \ie \textit{source graph}, to build a prompted graph, \ie \textit{target graph}.
GPL follows three steps~\cite{sun2023all}:
\textit{(1)~Graph prompting}: A graph prompting function $f_{p}(\cdot|\cdot)$ is leveraged to transform a source graph $G$ into a target graph $\tilde{G}$ as $\tilde{G} = f_p(G|P_G)$ where $P_G$ is a specific prompt for $G$.
\textit{(2)~Graph representing}: The GNN $\mathcal{G}$ produces a representation $H$ of the target graph as $H = \mathcal{G}(\tilde{G})$.
\textit{(3)~Task prediction}: A task head $f_{h}(\cdot)$ inputs $H$ to predict the label $y$ of the downstream task as $y = f_h(H)$. Usually, the task head is learnable for more accurate downstream prediction. For example, the task head can be a linear classifier~\cite{sun2023all}.

\section{Methodology}

The sec will illustrate the framework, algorithms, and training.
Specifically, we first propose the FedGPL framework to splitly enable privacy-preserving and personalized graph prompt tuning.
Then, we develop two algorithms, including a hierarchical directed transfer aggregator and a virtual prompt graph, to eliminate task and data heterogeneity, respectively.
Last, we present the training workflow based on algorithms tailored for the FedGPL framework.

\subsection{Framework}\label{sec:framework}

For an FGL system with both task and data heterogeneity, learning a parameter-shared model for all participants as traditional federated learning is not applicable. Because it may cause poor generalization ability on different tasks and divergent data.
Along with this idea, we propose a framework, namely \textbf{Federated Graph Prompt Learning~(FedGPL)}, to splitly preserve global and local knowledge by prompt-based privacy-preserving fine-tuning, illustrated in Figure~\ref{fig:framework}.
In light of Split Federated Learning~\cite{thapa2022splitfed}, the common and domain-specific graph knowledge can be captured separately, allowing global generalizability and local personalization.

\textit{(a)~Architecture.}
Overall, FedGPL consists of a server and several clients corresponding to participants.
First, we follow the GPL formulated in Section~\ref{sec:gpl} to unify learning procedure of three levels of tasks.
Second, we deploy a trainable GNN in the server to deliver the graph representing ability for heterogeneous clients.
The GNN contains versatile graph knowledge, which can be efficiently fine-tuned to fit multiple domains in a federated system.
Third, a task head and a graph prompt are personally optimized within each client for domain-specific understanding of its task and data.

\textit{(b)~Collaborative learning procedure.}
Based on the GPL pipeline, we describe how to activate forward- and backward- propagation with collaboration between the server and clients via communication.
The communication messages between modules include four elements: featured prompted graphs and their representation in forward-propagation, and gradients of the task head and the graph prompt in back-propagation.
On the one hand, the forward-propagation procedure follows:
(1)~A client prompts an input graph and sends it to the server.
(2)~The server generates the representation of the received prompted graph by GNN and then sends it back to the client.
(3)~The client takes the prompted graph representation as input of the task head for prediction.
On the other hand, the back-propagation procedure follows:
(1)~A client computes gradients of the task head and sends them to the server.
(2)~The server computes gradients of the prompted graph and sends it back to the client.
(3)~The client computes the gradients of the graph prompt.
After two-sided propagation, GPL modules can be optimized.

\textit{(c)~Differentially privacy preservation.}
The communication messages remain privacy leakage risks. Specifically, the server may obtain the raw features of prompted graphs from clients by attacks such as DLG~\cite{zhu2019deep}. Therefore, to further protect the client-side data privacy, we utilize a Differential Privacy~(DP) technique, Laplacian noise, to add it to embeddings and gradients as
\begin{equation}\label{equ:laplace}
    f(X; \epsilon) = X + \eta, \eta \sim \frac{1}{\lambda } \mathrm{exp} (-\frac{\left | x \right | }{\lambda } ), \lambda = \frac{\mathcal{V}(X)}{\epsilon } ,
\end{equation}
where $X$ is the embedding or gradient to be protected, $\mathcal{V}(X)$ determines the variation of the protected data, and $\epsilon$ is the privacy scale parameter.

In summary, based on the split strategy and DP techniques, we build a privacy-preserving federated framework to allow GPL to prevent private data from being leaked or inferred. 
Besides, state-of-the-art pre-trained GNNs such as GROVER~\cite{rong2020self} and DVMP~\cite{zhu2023dual} have approximately $100$ million parameters. Our split strategy makes clients without heavy memory burden to store GNNs, especially when applying FedGPL among resource-constrained devices~\cite{chen2019deep}.

\subsection{Algorithms}\label{sec:modules}

The existing federated algorithms~\cite{li2020federated,karimireddy2020scaffold} developed on FedAvg~\cite{mcmahan2017communication} can only be applied in traditional non-IID settings. However, they still fail in FGL with task and data heterogeneity. Because these algorithms strictly require a consistent model structure across clients. The requirement is infeasible in a task-heterogeneous federation that is more general in FGL. How to tackle the joint heterogeneity in this scenario for more effective federated optimization remains a significant difficulty.

Therefore, we propose an aggregating and a graph prompting algorithms to eliminate damage from graph task and data heterogeneity by a privacy-preserving and collaborative optimization.
Specifically, we first devise a federated aggregation algorithm in the server side to transfer asymmetrically beneficial knowledge between different tasks. Moreover, we equip each client with an graph prompt to retain useful local knowledge with less data divergence.

\subsubsection{Server-side federated aggregation.}\label{sec:hidra}

Previous methods such as FedAvg mostly aggregate local models based on an assumption that the heterogeneity between clients affects symmetrically. However, cross-task knowledge transfer is asymmetric, which means that the effectiveness of transferring a task to another may differ from the inverse. In fact, measuring the optimal directed knowledge transfer and activating it with learning benefits are both non-trivial problems in task-heterogeneous FGL.
Therefore, we develop the \textbf{Hierarchical Directed Transfer Aggregator~(HiDTA)} to federally enhance prediction performance of task-heterogeneous clients via directional knowledge transfer that is hierarchically determined by a proposed metric for transferring availability.

\textit{(1)~Transferability Definition: a metric for knowledge transfer availability.}
First, we define a metric, \ie transferability, to model the directed heterogeneity between clients in Definition~\ref{def:transferability}.
For a source client and a target client from their corresponding clients, transferability reflects \textit{how much benefit of the target client if it uses the source client's model on the target client's data}.
\begin{definition}
    \label{def:transferability}
    \textbf{Transferability.} We define the transferability $\tau_{a \gets b}$ from model $a$ to model $b$ as
    \begin{equation}\label{equ:trans_1}
        \tau_{a \gets b}^{(l)} = \frac{\overrightarrow{\theta_a^{(l)\prime}-\theta_a^{(l)}}\cdot\overrightarrow{\theta_b^{(l)\prime}-\theta_a^{(l)}}}{\Vert\overrightarrow{\theta_a^{(l)\prime}-\theta_a^{(l)}}\Vert},
    \end{equation}
    where $\theta_a$ is the parameters of $a$ model, $\theta^{(l)}$ is the parameters at $l$-th optimization step, $\theta^{(l)\prime}$ is the estimated optimal parameters computed on local data for $\theta^{(l)}$ at the next step.
\end{definition}
Transferability fits the property that knowledge transferring is often asymmetric, \ie $\tau_{i \gets j} \neq \tau_{j \gets i}$, which means that it is more convincing to measure the transferability by a directed value.

\textit{(2)~Hierarchical transferability evaluation.}
Next, we propose to evaluate pairwise transferability to understand the task heterogeneity based on the decoupled hierarchy of clients.
Specifically, we construct task-models to identify task-specific knowledge by fused perception on corresponding models.
For each task, we measure the multi-dimensional distances of parameters between client-models and task-models.
Based on task-models and calculated distances, we extrapolate the transferability for any two client-models by
\begin{equation}\label{equ:trans_2}
    \tau_{a \gets b}^{(l)} = (1-\sigma(\Vert\theta_a^{(l)}-\theta_{p_a}^{(l)}\Vert))\cdot(1-\sigma(\Vert\theta_b^{(l)}-\theta_{p_b}^{(l)}\Vert))\cdot\tau_{p_a \gets p_b}^{(l)},
\end{equation}
where $a$ and $b$ are two client-models, $p_a$ and $p_b$ are $a$ and $b$'s task-models, and $\sigma(\cdot)$ is a normalization function~(\eg $\operatorname{Sigmoid}(\cdot)$ or $\operatorname{ReLU}(\cdot)$).

\textit{(3)~Asymmetric knowledge transfer via model aggregation.}
Last, we weighted aggregate models to transfer knowledge from source clients to target clients based on normalized transferability values. The model parameters of a client $i$ can be compute as
\begin{equation}\label{equ:trans_update}
    \theta_i^{(l+1)}=\frac{\sum^{j=1}_{N}{(\tau_{i \gets j}^{(l)}\cdot\theta_j^{(l)})}}{\sum^{j=1}_{N}{\tau_{i \gets j}^{(l)}}},
\end{equation}
where $\theta_i^{(l+1)} = (\theta_{p,i}^{(l+1)}, \theta_{h,i}^{(l+1)})$ is the parameter set of $f_p(\cdot)$ and $f_h(\cdot|\cdot)$ of client $i$, and $N$ is the number of participating clients.

In short, HiDTA assists positive knowledge transfer across clients and impedes harmful attacks aroused by task heterogeneity, which are both extracted based on the hierarchical transferability assessment. The success of tackling task heterogeneity will be theoretically analyzed in Section~\ref{sec:task_het}.

\subsubsection{Client-side graph prompting.}\label{sec:vpg}
Existing works have proposed to construct universal graph prompts for any level of tasks by incorporating task-specific properties~\cite{sun2023all,fang2022prompt,zhu2023sgl}. However, there is an underexplored conflict between distilling specific knowledge and eliminating corresponding data heterogeneity. On the one hand, prompting with more task-specific information on graphs brings more severe data heterogeneity. On another hand, extracting more unique patterns requires more distinguished prompted features.
Besides, existing graph prompting strategies remain inefficient due to heavy models carrying prior knowledge, which is unsuitable for large-scale FGL scenarios.
Therefore, we devise \textbf{Virtual Prompt Graph~(VPG)}, to model task-specific knowledge as well as minimize data heterogeneity by adaptively indicating intellective graph structures.

\textit{(1)~Prompting module construction.}
First, we insert a positive structure to capture distinctive information in the particular domain.
Second, we construct a negative structure to neutralize noisy nodes and edges in terms of relatively useful graph structures.
Simultaneously, VPG can assist the GNN to generate more effective yet less heterogeneous graph representations for tailored clients.
Then, we define VPG in Definition~\ref{def:vpg}, which implies how we can transform a \textit{source graph}, \ie the original graph, into a \textit{target graph}, \ie the prompted graph, to be input into the GNN.

\begin{definition}\label{def:vpg}
    \textbf{VPG.} For a graph $G=(V,E,X)$, we denote a VPG as $\hat{G}=((\hat{V}^+,\hat{V}^-),(\hat{E}^+,\hat{E}^-),(\hat{X}^+,\hat{X}^-))$. It is satisfied that $\hat{V}^+ \cap \hat{V}^- = \emptyset$, where $\hat{V}^+=\{v_{N_n+1}^+,v_{N_n+2}^+,\cdots\}$ is the prompted node set and $\hat{V}^-=\{v_{i_1}^-,v_{i_2}^-,\cdots\}, v_{i_1},v_{i_2},\cdots \in V$ is the prompted anti-node set.
    And it is satisfied that $\hat{E}^+ \cap \hat{E}^- = \emptyset$, where $\hat{E}^+=\{e_{N_e+1}^+,e_{N_e+2}^+,\cdots\}$ is the prompted edge set and $\hat{E}^-=\{e_{j_1}^-,e_{j_2}^-,\cdots\}, e_{j_1},e_{j_2},\cdots \in E$ is the prompted anti-edge set.
\end{definition}

\textit{(2)~Learning for adaptive prompting on graphs.}
We define VPG-based prompt function as $\tilde{G} = f_p(G|\hat{G})$, where $f_p(\cdot|\cdot)$ follows $\tilde{V} = V \cup \hat{V}^+ \setminus \hat{V}^-$, $\tilde{E} = E \cup \hat{E}^+ \setminus \hat{E}^-$, $\tilde{X} = \{x_i|v_i \in \tilde{V}\}$.
Then, we propose a generative algorithm to build specific VPGs.
Specifically, we construct a candidate set of $k^{\prime}$ nodes, whose features are pre-defined~(default) or learnable. We connect them to the source graph if $v_i$ and $v_j$ satisfies $\operatorname{Sigmoid}(x_i \cdot x_j) < \gamma$, where $\gamma$ is a threshold. After that, we have $G^{\prime}=(V^{\prime},E^{\prime},X^{\prime})$.
We generate node significance as $w^{\prime}_n = X^{\prime}\cdot p^{\prime}/||p^{\prime}||$, where $p^{\prime} \in \mathbb{R}^{d}$ is a learnable vector.
We also generate edge significance between $i$-th and $j$-th node as $w^{\prime}_{e,i,j}=||w^{\prime}_{n,i}-w^{\prime}_{n,j}||$.
The node features will be also processed by $x_i=x_i\cdot \operatorname{tanh}(w^{\prime}_{n,i})$.
Next, we determine the node significance borderline $\bar{w}^{\prime}_n$ as the $k_n$-th highest significance score among $V^{\prime}$, and the edge significance borderline $\bar{w}^{\prime}_e$ in a similar way. Note that $k_n=\alpha_n\cdot |V|$ and $k_e=\alpha_e\cdot |E|$ and we manually decide $\alpha_n$ and $\alpha_e$.
Last, we can create a VPG $\hat{G}$ by constructing the prompted (anti-)~node set and (anti-)~edge set, by letting $\hat{V}^+ = V^{\prime}_{w^{\prime}_{n,i} \geq \bar{w}^{\prime}_n} \setminus V$, $\hat{V}^- = V_{w^{\prime}_{n,i} < \bar{w}^{\prime}_n}$, $\hat{E}^+ = E^{\prime}_{w^{\prime}_{e,i,j} \geq \bar{w}^{\prime}_e,w^{\prime}_{n,i} \geq \bar{w}^{\prime}_n,w^{\prime}_{n,j} \geq \bar{w}^{\prime}_n} \setminus E$, and $\hat{E}^- = E_{w^{\prime}_{e,i,j} < \bar{w}^{\prime}_e|w^{\prime}_{n,i} < \bar{w}^{\prime}_n|w^{\prime}_{n,j} < \bar{w}^{\prime}_n}$.

In conclusion, VPG indicates useful and useless graph structures to simplify local graph data to reduce data heterogeneity with better representation production, which will be discussed in Section~\ref{sec:data_het}.

\subsection{Training}\label{sec:training}
We present the training workflow of FedGPL.
First, we allow clients to independently optimize their kept modules including prompting and task head modules.
Second, the server-side GNN can be optionally optimized. It can be pre-trained or randomly initialized. After that, its parameters will be fine-tuned or frozen during optimization.
In this work, we choose to fine-tune a pre-trained GNN by default, because pre-training and fine-tuning GNNs can efficiently utilize common graph knowledge and fit on specific domains.

Detailedly, we take the node classification task as an example in Appendix~\ref{sec:appendix_training}.
Specifically, we first initialize the client-side parameters.
For each training round, we enable each client in parallel to individually train their modules. The client cooperates with the server for split and privatized forward- and back- propagation.
Then the server globally calculates transferability between clients and coordinates them to locally update parameters.
The training process will terminate after parameters in clients are converged.

\begin{table*}[t]
\setlength{\tabcolsep}{6pt}
\caption{ACC~($\%$) and F1~($\%$) of different federated algorithms and graph prompting methods. We fine-tune all GPL modules including a pre-trained GraphTransformer~\cite{yun2019graph}, graph prompts, and task heads.}
\label{tab:overall}
\begin{tabular}{cc|cc|>{\columncolor[HTML]{EFEFEF}}c 
>{\columncolor[HTML]{EFEFEF}}c |cc|>{\columncolor[HTML]{EFEFEF}}c 
>{\columncolor[HTML]{EFEFEF}}c |cc}
\hline
\multicolumn{1}{c|}{}                       &                                   & \multicolumn{2}{c|}{Cora}    & \multicolumn{2}{c|}{\cellcolor[HTML]{EFEFEF}CiteSeer} & \multicolumn{2}{c|}{DBLP}    & \multicolumn{2}{c|}{\cellcolor[HTML]{EFEFEF}Photo} & \multicolumn{2}{c}{Physics}  \\
\multicolumn{1}{c|}{\multirow{-2}{*}{\begin{tabular}[c]{@{}c@{}}Federated\\ Method\end{tabular}}} & \multirow{-2}{*}{\begin{tabular}[c]{@{}c@{}}Prompt\\ Method\end{tabular}} & ACC  & F1   & ACC   & F1   & ACC  & F1   & ACC   & F1   & ACC  & F1   \\ \hline
\multicolumn{1}{c|}{}                       & GPF                               & 85.02 & 84.37 & 84.60 & 84.18 & 81.03 & 80.70 & 76.85 & 76.46 & 88.29 & 88.49 \\
\multicolumn{1}{c|}{}                       & ProG                              & 86.45 & 85.91 & 85.72 & 85.85 & 82.20 & 82.01 & 71.47 & 70.47 & 86.73 & 86.98 \\
\multicolumn{1}{c|}{\multirow{-3}{*}{Local}} & SUPT                              & 85.66 & 85.23 & 87.00 & 86.88 & 80.57 & 80.52 & 73.31 & 72.87 & 89.69 & 89.56 \\ \hline
\multicolumn{1}{c|}{}                       & GPF                               & 85.20 & 84.76 & 85.36 & 85.22 & 81.87 & 81.88 & 80.73 & 80.52 & 89.75 & 88.83 \\
\multicolumn{1}{c|}{}                       & ProG                              & 86.70 & 86.13 & 86.49 & 86.05 & 82.43 & 82.21 & 72.21 & 71.76 & 89.05 & 89.27 \\
\multicolumn{1}{c|}{\multirow{-3}{*}{FedAvg}} & SUPT                              & 85.74 & 85.23 & 87.17 & 84.05 & 83.07 & 82.52 & 81.70 & 81.44 & 89.79 & 88.60 \\ \hline
\multicolumn{1}{c|}{}                       & GPF                               & 80.52 & 79.74 & 83.20 & 82.69 & 80.57 & 79.84 & 82.93 & 82.79 & 87.60 & 87.19 \\
\multicolumn{1}{c|}{}                       & ProG                              & 84.04 & 83.79 & 84.39 & 83.87 & 81.86 & 81.55 & 72.45 & 72.46 & 89.01 & 88.57 \\
\multicolumn{1}{c|}{\multirow{-3}{*}{FedProx}} & SUPT                              & 81.81 & 80.59 & 83.47 & 83.82 & 80.79 & 80.80 & 80.26 & 80.64 & 89.21 & 88.21 \\ \hline
\multicolumn{1}{c|}{}                       & GPF                               & 77.08 & 76.39 & 75.79 & 74.87 & 78.57 & 77.63 & 74.43 & 74.37 & 85.20 & 84.85 \\
\multicolumn{1}{c|}{}                       & ProG                              & 73.90 & 73.25 & 74.80 & 74.57 & 76.99 & 76.25 & 72.36 & 71.72 & 85.60 & 85.27 \\
\multicolumn{1}{c|}{\multirow{-3}{*}{SCAFFOLD}} & SUPT                              & 75.38 & 70.47 & 74.91 & 73.73 & 75.80 & 75.71 & 75.65 & 75.86 & 85.53 & 84.49 \\ \hline
\multicolumn{2}{c|}{\textbf{FedGPL}}                                                         & \textbf{88.34} & \textbf{88.07} & \textbf{87.60} & \textbf{87.72} & \textbf{83.93} & \textbf{83.27} & \textbf{86.75} & \textbf{86.03} & \textbf{90.91} & \textbf{90.61} \\ \hline
\end{tabular}

\end{table*}

\section{Theoretical Analysis}

This part investigates FedGPL about (1)~the reduction of task heterogeneity, (2)~the elimination of data heterogeneity, and (3) the complexity of memory and communication.

\subsection{Task Heterogeneity Analysis}\label{sec:task_het}

We analyze the effect of our aggregation algorithm, HiDTA, on task heterogeneity by quantifying the heterogeneity values based on Definition~\ref{def:task_het}.
We theoretically propose Theorem~\ref{the:task_het} to demonstrate the success of HiDTA in eliminating task heterogeneity, which is proved in Appendix~\ref{sec:appendix_proof_task_het}.

\begin{theorem}\label{the:task_het}
    For $a$ -th and $b$ -th clients in FedGPL, we denote their estimated optimized parameters at $l$-th step as $\theta_a^{(l+1)\prime},\theta_b^{(l+1)\prime}$~(aggregated) or $\theta_a^{(l)\prime},\theta_b^{(l)\prime}$~(non-aggregated). It is satisfied that
    \begin{equation}
        \Delta_T^{a,b}(\theta_a^{(l+1)\prime},\theta_b^{(l+1)\prime}) \leq \Delta_T^{a,b}(\theta_a^{(l)\prime},\theta_b^{(l)\prime}),
    \end{equation}
    when their bidirectional transferability are positive. Hence, the task heterogeneity can be reduced by HiDTA.
\end{theorem}

\subsection{Data Heterogeneity Analysis}\label{sec:data_het}

We quantify data heterogeneity influenced by VPG based on Definition~\ref{def:data_het}. We compute the theoretical values before and after prompting and propose Theorem~\ref{the:data_het} which can be proved by Appendix~\ref{sec:appendix_proof_data_het}.

\begin{theorem}\label{the:data_het}
    For the $a$ -th and $b$ -th clients in FedGPL, graph representations from two clients are $h_{\tilde{G}_a},h_{\tilde{G}_b}$ and $h_{\tilde{G}_a},h_{\tilde{G}_b}$ for with and without prompting methods, respectively. And their graph data following uniform distributions of $h_{G_a} \sim U^a[0, \eta^a]$ and $h_{G_b} \sim U^b[0, \eta^b]$, respectively. We ensure
    \begin{equation}
        \mathbb{E}_{\left\{\substack{h_{\tilde{G}_a} \sim \tilde{U}^a\\h_{\tilde{G}_b} \sim \tilde{U}^b}\right.}[\Delta_{D}^{i,j}(h_{\tilde{G}_a},h_{\tilde{G}_b})] \leq \mathbb{E}_{\left\{\substack{h_{G_a} \sim U^a\\h_{G_b} \sim U^b}\right.}[\Delta_{D}^{i,j}(h_{G_a},h_{G_b})],
    \end{equation}
    when $\alpha_n^a \approx \alpha_n^b$.
    Thus, the data heterogeneity can be reduced by VPG.
\end{theorem}

\subsection{Efficiency Analysis}

\subsubsection{Memory efficiency.}
We assume the input graph has $N$ nodes and $M$ edges, the GNN has $L$ layers with the maximum dimension of a layer being $h$, and the task head has $l$ layers.
Here we compute the theoretical storage complexity.
The client has to store the whole GNN in the usual federated prompting setting. Hence, we take a popular GNN variant, GAT~\cite{velivckovic2017graph}, as an example of the GNN, the storage complexity will be $O(LKh^2+LKh+Kh)$. By applying our split strategy, each client only needs to store the prompt and a task head layer, where the storage complexity is $O(h+lh)$.

\subsubsection{Communication efficiency.}
The communication includes two parts in FedGPL that are between clients and the server~(depending on the graph and embedding sizes), and between clients and the server~(depending on the prompt and task head sizes).
Theoretically, VPG appends $k^{\prime}$ candidate nodes into the input graph, then score by a learnable vector $p^{\prime}$, the final prompted graph contains $k_{n}$ nodes, as the candidate nodes are pre-defined, the parameter complexity is $O(h)$, if we set the candidate nodes as trainable, the parameter complexity is $O(k^{\prime}h+h)$. Thus, VPG contains minimum $d$ parameters.
In addition, we involve two baseline GPL methods for comparison, including GPF~(parameters:~$O(nh)$) and ProG~(parameters:~$O(kh)$).

\section{Experiments}

We conduct experiments to evaluate our methods by answering the following questions in particular:
\textbf{Q1}: How effective is the overall prediction performance of FedGPL and each module?
\textbf{Q2}: How useful is FedGPL against task and data heterogeneity, respectively?
\textbf{Q3}: How efficient is our method in terms of memory, communication, and time on large-scale data?
Our codes, data, and appendix are open in \url{https://anonymous.4open.science/r/FedGPL/}.

\begin{table*}[t]
\centering
\caption{ACC~($\%$) of federated algorithms under different task combinations on Cora. Tasks in the federation are \colorbox{mycolor}{blue}. N, E, and G:~node-, edge-, and graph-level task. N+E: federation between node- and edge- level tasks. Global: the average of three tasks.}
\setlength{\tabcolsep}{2pt}
\begin{tabular}{@{}c|cccc|cccc|cccc@{}}
\toprule
                         & \multicolumn{4}{c|}{N+E}                                                                                                                                                               & \multicolumn{4}{c|}{N+G}                                                                                                                                       & \multicolumn{4}{c}{E+G}                                                                                                                                        \\
\multirow{-2}{*}{Method} & N                                      & E                                      & \multicolumn{1}{c|}{G}                                      & \cellcolor[HTML]{FFFFFF}Global         & N                                      & E                                      & \multicolumn{1}{c|}{G}                                      & Global         & N                                      & E                                      & \multicolumn{1}{c|}{G}                                      & Global         \\ \midrule
FedAvg                   & \cellcolor[HTML]{E0E7F9}$79.02$          & \cellcolor[HTML]{E0E7F9}$89.34$          & \multicolumn{1}{c|}{$91.02$}                                  & \cellcolor[HTML]{FFFFFF}$86.46 $         & \cellcolor[HTML]{E0E7F9}$79.09$          &$ 90.35 $                                 & \multicolumn{1}{c|}{\cellcolor[HTML]{E0E7F9}$90.02$}          & $86.49$          & $79.19   $                               & \cellcolor[HTML]{E0E7F9}$90.13$          & \multicolumn{1}{c|}{\cellcolor[HTML]{E0E7F9}$91.14$}          & $86.82 $         \\
FedProx                  & \cellcolor[HTML]{E0E7F9}$73.96$          & \cellcolor[HTML]{E0E7F9}$89.58$          & \multicolumn{1}{c|}{$91.02$}                                  & \cellcolor[HTML]{FFFFFF}$84.85$          & \cellcolor[HTML]{E0E7F9}$79.15$          &$ 90.35  $                                & \multicolumn{1}{c|}{\cellcolor[HTML]{E0E7F9}$91.11$}          & $86.87$          & $79.19$                                  & \cellcolor[HTML]{E0E7F9}$91.52$          & \multicolumn{1}{c|}{\cellcolor[HTML]{E0E7F9}$91.14$}          & $87.28$          \\
SCAFFOLD                 & \cellcolor[HTML]{E0E7F9}$71.21$          & \cellcolor[HTML]{E0E7F9}$89.67 $         & \multicolumn{1}{c|}{$91.02$}                                  & \cellcolor[HTML]{FFFFFF}$83.97$          & \cellcolor[HTML]{E0E7F9}$72.53$          & $90.35$                                  & \multicolumn{1}{c|}{\cellcolor[HTML]{E0E7F9}$80.33$}          &$ 81.07$          & $79.19$                                  & \cellcolor[HTML]{E0E7F9}$90.21 $         & \multicolumn{1}{c|}{\cellcolor[HTML]{E0E7F9}$80.68$}          & $83.36$          \\
\midrule
& \cellcolor[HTML]{E0E7F9}$79.73 $         & \cellcolor[HTML]{E0E7F9}$90.89 $         & \multicolumn{1}{c|}{$91.02$}                                  & \cellcolor[HTML]{FFFFFF}$87.21$          & \cellcolor[HTML]{E0E7F9}$80.44$          & $90.35$                                  & \multicolumn{1}{c|}{\cellcolor[HTML]{E0E7F9}\textbf{92.56}} & $87.78$          & $79.19 $                                 & \cellcolor[HTML]{E0E7F9}\textbf{92.13} & \multicolumn{1}{c|}{\cellcolor[HTML]{E0E7F9}\textbf{92.75}} & $88.02$          \\
\multirow{-2}{*}{HiDTA}                    & \cellcolor[HTML]{E0E7F9}\textbf{80.80} & \cellcolor[HTML]{E0E7F9}\textbf{91.76} & \multicolumn{1}{c|}{\cellcolor[HTML]{E0E7F9}\textbf{92.12}} & \cellcolor[HTML]{FFFFFF}\textbf{88.34} & \cellcolor[HTML]{E0E7F9}\textbf{80.80} & \cellcolor[HTML]{E0E7F9}\textbf{91.76} & \multicolumn{1}{c|}{\cellcolor[HTML]{E0E7F9}$92.12$}          & \textbf{88.34} & \cellcolor[HTML]{E0E7F9}\textbf{80.80} & \cellcolor[HTML]{E0E7F9}$91.76$          & \multicolumn{1}{c|}{\cellcolor[HTML]{E0E7F9}$92.12$}          & \textbf{88.34} \\ \bottomrule
\end{tabular}
\label{tab:task_het}
\end{table*}

\subsection{Experimental Setup}

\subsubsection{Datasets.}
We conduct experiments on five datasets including Cora~\cite{chen2021fedgraph}, CiteSeer~\cite{chen2021fedgraph}, DBLP~\cite{bojchevski2017deep}, Photo~\cite{shchur2018pitfalls}, and Physics~\cite{shchur2018pitfalls}. We follow \cite{sun2023all} for constructing datasets of node-, edge- and graph- level tasks.
Each data sample is a graph with about $300$ nodes. Usually, we can generate more than $400$ samples for each client.
We deploy multiple clients~($3$ by default) for each level of task in the following experiments, all clients participate in each round of training, and each client conducts local training once per round.
The detailed statistic of the original graph datasets is listed in Appendix~\ref{sec:appendix_dataset}.

\subsubsection{Baselines.}
We combine the state-of-the-art GPL and FL methods to construct competitive baselines. First, we select three GPL methods available for all three levels of graph tasks, including (1)~ProG~\cite{sun2023all}, which insert an extra graph as a prompt into the input graph, (2)~GPF~\cite{fang2022prompt}, which inserts extra vectors as prompts into the nodes' feature, and (3)~SUPT~\cite{lee2024subgraph}, which prompt features in the subgraph.
Second, we select three FL methods, including FedAvg~\cite{mcmahan2017communication}, FedProx~\cite{li2020federated}, and SCAFFOLD~\cite{karimireddy2020scaffold} to aggregate corresponding GPL modules.

\subsubsection{Evaluation metrics.}
We use accuracy~(ACC) and F1 Score~(F1) to evaluate the prediction performance of tested methods on the node classification task.


\begin{table*}[t]
\setlength{\tabcolsep}{3pt}
\caption{Performance of different federated algorithms against data heterogeneity. Minor $\alpha$ means higher data heterogeneity.}
\begin{tabular}{@{}c|cccc|cccc|cccc|cccc@{}}
\toprule
\multirow{3}{*}{\begin{tabular}[c]{@{}c@{}}Federated\\ Method\end{tabular}} & \multicolumn{4}{c|}{$\alpha = 0.9$}                                                               & \multicolumn{4}{c|}{$\alpha = 0.6$}                                                               & \multicolumn{4}{c|}{$\alpha = 0.3$}                                                               & \multicolumn{4}{c}{$\alpha = 0.1$}                                                                \\ \cmidrule(l){2-17} 
                                                                            & \multicolumn{2}{c|}{Cora}                            & \multicolumn{2}{c|}{CiteSeer}   & \multicolumn{2}{c|}{Cora}                            & \multicolumn{2}{c|}{CiteSeer}   & \multicolumn{2}{c|}{Cora}                            & \multicolumn{2}{c|}{CiteSeer}   & \multicolumn{2}{c|}{Cora}                            & \multicolumn{2}{c}{CiteSeer}    \\
                                                                            & ACC            & \multicolumn{1}{c|}{F1}             & ACC            & F1             & ACC            & \multicolumn{1}{c|}{F1}             & ACC            & F1             & ACC            & \multicolumn{1}{c|}{F1}             & ACC            & F1             & ACC            & \multicolumn{1}{c|}{F1}             & ACC            & F1             \\ \midrule
FedAvg                                                                      & 86.93          & \multicolumn{1}{c|}{86.40}          & 86.93          & 86.42          & 87.77          & \multicolumn{1}{c|}{87.02}          & 86.28          & 85.87          & 85.00          & \multicolumn{1}{c|}{84.59}          & 83.87          & 83.62          & 83.43          & \multicolumn{1}{c|}{83.39}          & 82.13          & 81.92          \\
FedProx                                                                     & 85.00          & \multicolumn{1}{c|}{84.45}          & 85.71          & 85.57          & 84.07          & \multicolumn{1}{c|}{83.63}          & 85.77          & 85.29          & 82.43          & \multicolumn{1}{c|}{82.28}          & 83.09          & 82.90          & 81.90          & \multicolumn{1}{c|}{81.35}          & 81.57          & 81.20          \\
SCAFFOLD                                                                    & 71.67          & \multicolumn{1}{c|}{71.87}          & 73.07          & 72.59          & 74.17          & \multicolumn{1}{c|}{73.60}          & 73.07          & 72.60          & 71.65          & \multicolumn{1}{c|}{71.90}          & 73.12          & 73.07          & 72.87          & \multicolumn{1}{c|}{72.28}          & 71.94          & 70.64          \\ \midrule
FedGPL                                                                       & \textbf{88.23} & \multicolumn{1}{c|}{\textbf{87.83}} & \textbf{87.33} & \textbf{87.39} & \textbf{87.90} & \multicolumn{1}{c|}{\textbf{87.42}} & \textbf{86.93} & \textbf{86.69} & \textbf{85.07} & \multicolumn{1}{c|}{\textbf{84.07}} & \textbf{82.77} & \textbf{82.60} & \textbf{83.93} & \multicolumn{1}{c|}{\textbf{83.72}} & \textbf{82.29} & \textbf{82.23} \\ \bottomrule
\end{tabular}
\label{tab:data_het}
\end{table*}

\subsection{RQ1: Federated Prompt Tuning Performance}

\subsubsection{Overall results.}
Here we compare FedGPL with baselines by averaging the performance of clients. Their performance is evaluated according to prediction ACC and F1 on node classification tasks on five datasets.
First, we train all modules in these methods, including the pre-trained GNN, graph prompts, and task heads. As shown in Table~\ref{tab:overall}, FedGPL consistently outperforms all other algorithms, achieving an improvement range of $2.37\%$ to $16.07\%$ over state-of-the-art methods by jointly incorporating HiDTA and VPG to tackle multifaceted heterogeneity.

\begin{figure}[t]
\setlength{\abovecaptionskip}{0.1cm}
\setlength{\belowcaptionskip}{-0.3cm}
    \centering
    \subfigure[ACC]{
        \includegraphics[width=0.46\linewidth]{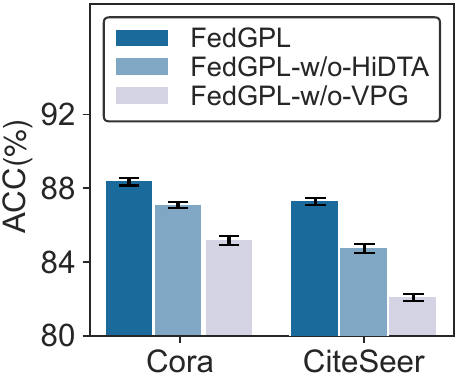}
        \label{fig:ablation_acc_1}
    }
    \subfigure[F1]{
        \includegraphics[width=0.46\linewidth]{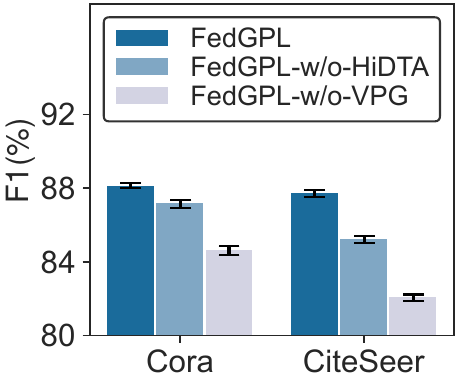}
        \label{fig:ablation_f1_1}
    }
    \caption{Ablation study.}
    \label{fig:ablation}
\end{figure}

\subsubsection{Ablation study.}
To evaluate the effectiveness of FedGPL modules, we conduct an ablation study with two variations. \textit{FedGPL-w/o-HiDTA} refers to a variant that excludes the HiDTA module, while \textit{FedGPL-w/o-VPG} denotes another variant without the VPG module.
Figure~\ref{fig:ablation} shows significant performance drops when HiDTA or VPG is removed, confirming their effectiveness in our framework. Specifically, the ACC of FedGPL-w/o-HiDTA decreases by $1.04\%$ on Cora, and by $1.55\%$ on CiteSeer. We explain that HiDTA facilitates adaptive aggregation that enhances cross-client knowledge sharing, resulting in performance improvements compared to isolated training.
Additionally, ACC of FedGPL-w/o-VPG decreases by $3.31\%$ on Cora, and $3.62\%$ on CiteSeer, indicating that VPG allows transformed input graphs to be well understood by the GNN. We also test the statistical significance of prediction performance by varying random seeds. For example, for FedGPL, ACC is $0.8817 \pm 0.1650$ and F1 is $0.8772 \pm 0.1672$ on Cora.

\subsection{RQ2: Effectiveness of Tackling Graph Task and Data Heterogeneity}

We evaluate the effectiveness against heterogeneous graph tasks and data under the FedGPL framework according to the prediction ACC tested on varied settings.

\subsubsection{Task heterogeneity.}
We construct three settings with heterogeneous tasks, which only involve two tasks in federated optimization and another one is trained locally.
We present the results in Table~\ref{tab:task_het}. On the one hand, compared with baseline aggregation algorithms, HiDTA performs better under multiple task-heterogeneous scenarios. For example, the ACC of HiDTA improves at least $1.63\%$ in node-level tasks and $1.59\%$ in graph-level tasks when they are federally optimized. On the other hand, HiDTA achieves a higher average ACC with a federation among three tasks than two tasks. For example, the three-task federation gains $1.10\%$ enhancement compared with a federation between node-level tasks and edge-level tasks. Besides, a special case is that edge-level tasks and graph-level tasks perform slightly better in their federation than the ones involving node-level tasks. On the contrary, node-level tasks obtain $1.97\%$ improvement through participation.
We observe a positive trade-off between tasks that although two levels of tasks drop their ACC, they significantly enhance node-level ones to reach a better overall ACC. The finding shows that HiDTA optimizes models by centering the global performance.

\subsubsection{Data heterogeneity.}
We vary the degree of data heterogeneity by following an existing work~\cite{yurochkin2019bayesian} to simulate non-IID scenarios with a $\alpha$-Dirichlet distribution. As $\alpha$ decreases, clients' data becomes more non-IID. We report results in Table~\ref{tab:data_het}. When graph data tend to be divergent, the ACC is reduced. In comparison, FedGPL outperforms baselines under different non-IID degrees. For example, on Cora, FedGPL improves over $1.35\%$, $0.38\%$, $0.35\%$ F1 of baselines when $\alpha=0.9$, $0.6$, and $0.3$, respectively. We conclude that VPG can eliminate extensive graph data heterogeneity by highlighting important graph structures while reducing useless subgraphs.

\begin{table}[t]
    \caption{Training time~($200$ epochs) and memory cost~(GPU memory usage per client) on the Physics dataset~($\sim120,000$ nodes per client) with $9$ clients.}
    \label{tab:time_memory}
    \begin{tabular}{@{}cc|cc@{}}
    \toprule
    \multicolumn{1}{c|}{\begin{tabular}[c]{@{}c@{}}Federated\\ Method\end{tabular}} & \multicolumn{1}{c|}{\begin{tabular}[c]{@{}c@{}}Prompt\\ Method\end{tabular}}  & Time~(min)          & Memory~(G)        \\ \midrule
    \multicolumn{1}{c|}{\multirow{3}{*}{FedAvg}}   & GPF                                                     & 9.52          & 2.29          \\
    \multicolumn{1}{c|}{}                          & ProG                                                    & 10.83         & 2.37          \\
    \multicolumn{1}{c|}{}                          & SUPT                                                    & 10.11         & 2.52          \\ \midrule
    \multicolumn{1}{c|}{\multirow{3}{*}{FedProx}}  & GPF                                                     & 9.84          & 2.32          \\
    \multicolumn{1}{c|}{}                          & ProG                                                    & 11.21         & 2.38          \\
    \multicolumn{1}{c|}{}                          & SUPT                                                    & 10.62         & 2.54          \\ \midrule
    \multicolumn{1}{c|}{\multirow{3}{*}{SCAFFOLD}} & GPF                                                     & 11.81         & 2.39          \\
    \multicolumn{1}{c|}{}                          & ProG                                                    & 14.61         & 2.57          \\
    \multicolumn{1}{c|}{}                          & SUPT                                                    & 12.63         & 2.48          \\ \midrule
    \multicolumn{2}{c|}{FedGPL}                                                                              & \textbf{7.51} & \textbf{0.43} \\ \bottomrule
    \end{tabular}
\end{table}

\begin{table}[t]
    \centering
    \caption{Evaluation of communication cost. $f_p(\cdot|\cdot)$:~prompting function; $\tilde{G}$:~prompted graph; Comm.:~communication.}
    \begin{tabular}{c|ccc}
    \toprule
    & $f_p(\cdot|\cdot)$ Size   & $\tilde{G}$ Size & Comm. Cost \\
    \midrule
    FedAvg+GPF       & $20,000$           & $20,000$                 & $81,600$              \\
    FedAvg+ProG      & $1,000$            & $21,000$                 & $45,600$              \\
    FedGPL      & \textbf{100} & \textbf{10,000}        & \textbf{21,800}     \\
    FedGPL*     & $1,100$          & \textbf{10,000}        & $23,800$              \\
    \bottomrule
    \end{tabular}
    \label{tab:communication}
\end{table}

\subsection{RQ3: Efficiency Evaluation}

\subsubsection{Memory efficiency.}
We calculate the number of stored parameters to evaluate the client-side memory burden. Specifically, our default setting demands $81,600$ parameters in each client. Our method requires each client only to store $800$ parameters, reducing $99.02\%$ memory space.
In practice, Table~\ref{tab:time_memory} shows that FedGPL achieves $5.3\times \sim 6.0\times$ GPU memory efficiency in the training environment compared to previous work, which implies the potential to adopt FedGPL on resource-constrained settings.

\subsubsection{Communication efficiency.}
Here we compute and compare the communication cost of four methods, including (1)~FedAvg+GPF, (2)~FedAvg+ProG, (3)~FedGPL, and (4)~FedGPL*: a variant of FedGPL incorporating a VPG with learnable $V^{+}$ features.
Experimental results are listed in Table~\ref{tab:communication}. Specifically, FedGPL and FedGPL* reduce over $70.83\%$ communication costs compared with FedAvg+GPF, and $47.80\%$ compared with FedAvg+ProG. The significant reduction is attributed to the smaller sizes of $f_p(\cdot|\cdot)$ and $\tilde{G}$. For $f_p(\cdot|\cdot)$, FedGPL requires only $0.5\%$ and $10\%$ parameters of GPF and ProG. The size of FedGPL's $\tilde{G}$ is $50\%$ and $47.62\%$ of FedAvg+GPF and FedAvg+ProG. In conclusion, VPG behaves $2.1\times \sim 3.7\times$ more communication-efficiently than baselines. Therefore, FedGPL is scalable for massive FGL participants with lightweight transmission between the server and clients.

\subsubsection{Time efficiency.}
We demonstrate the time efficiency for our method in Table~\ref{tab:time_memory}.
Specifically, FedGPL can decrease the training time cost from $21.1\%$ to $51.4\%$ compared to baseline methods.
The significant improvement in training time efficiency is attributed to reduced training parameters and communication messages. Thus, FedGPL is suitable for large-scale datasets with millions of nodes and edges in resource-constrained real-world FGL scenarios.

\section{Related Works}

Here we review works of two related research fields, including federated graph learning, and graph prompt learning.

\subsection{Federated Graph Learning}
Federated Learning~(FL)~\cite{mcmahan2017communication} is a distributed machine learning algorithm that allows multiple clients to collaborate without leakage of their data.
Studies have focused on various issues such as privacy-protection~\cite{wei2020federated,truex2019hybrid,zhang2023trading}, data heterogeneity~\cite{li2020federated, karimireddy2020scaffold,li2022federated}, model heterogeneity~\cite{tan2022fedproto,jang2022fedclassavg}, communication costs~\cite{konevcny2016federated,chen2021communication,almanifi2023communication}.
The integration of FL with graph learning, called Federated Graph Learning~(FGL), has provided promising solutions to many real-world problems~\cite{zhang2021federated,chen2021fedgraph,fu2022federated}.
Researchers study FGL to allow institutions to collaboratively learn models while keeping local graph data private~\cite{guo2024hifgl}.
Existing FGL works propose various GNN modules to improve downstream prediction effectiveness.
For example, FedSage+~\cite{zhang2021subgraph} generates local nodes' neighbor features to offset the ignorance of cross-client edges due to subgraph-level privacy preservation, which improves predicting ability with significant extra training costs.
Furthermore, recent works investigate to tackle the non-IID issue in FGL.
FedStar~\cite{tan2023federated} shares common structural knowledge across heterogeneous clients by a personalized module.
FedLIT~\cite{xie2023federated} automatically discovers link-type heterogeneity to alleviate the useless structure to enable more effective message passing.
FGGP~\cite{wan2024federated} addresses non-IID graphs by decoupling structure and attribute domain shift based on prototype learning.
However, they overlook task-heterogeneous FGL, a more general and difficult setting. Moreover, existing frameworks fail to adapt the scenario with heterogeneous downstream tasks.
Our work proposes a novel framework tailored for task- and data- heterogeneous FGL to fill this gap.

\subsection{Graph Prompt Learning}
Prompting methods have achieved significant success in Natural Language Processing~(NLP). Thus, researchers are exploring applying prompting methods to graph learning tasks. The heterogeneity among different graph learning tasks is more complex than NLP, which brings unique challenges.
A main branch of prompts for graph learning is inserting extra feature vectors as prompts into the feature space of nodes and graphs. 
\cite{fang2022prompt,gong2023prompt,liu2023graphprompt,zhu2023sgl,tan2023virtual} add trainable feature vectors to the node features, aligning different types of pre-training tasks and downstream tasks.
\cite{ma2023hetgpt} injects heterogeneous feature prompt into the input graph feature space.
Other methods design extra graphs as prompts and link them with the input graph. 
For example, \cite{sun2023all,ge2023enhancing,huang2023prodigy} design extra small graphs as prompts and link the prompt graph with the input graph to perform the prompting process.
The above works deploy prompting techniques on the features or structures of graph data.
In our work, we design VPG for featural and structural prompting, which adds extra knowledge by nodes and edges, and neutralizes noisy structures to reduce the harm from task and data heterogeneity.

\section{Conclusions}

This paper investigated the multifaceted heterogeneity of graph task and data in FGL, which is unexplored in previous studies. We first construct a split framework for the separate preservation of global and local knowledge. Moreover, we developed HiDTA, a federated aggregator to hierarchically share asymmetrically beneficial knowledge with task heterogeneity elimination. Next, we devise a graph prompt that is adaptively generated to emphasize informative structures and denoise useless ones to improve representation efficiency. We provide theoretical analysis to prove the successful mitigation of the graph task and data heterogeneity. We also conduct extensive experiments on our approach and baseline methods to show the accuracy and efficiency superiority of FedGPL on large-scale datasets with millions of nodes.



\clearpage
\bibliographystyle{ACM-Reference-Format}
\bibliography{references}


\clearpage
\appendix

\section{Appendix}

\subsection{Graph Learning Tasks}\label{sec:appendix_graph_task}

Here we define three levels of graph learning tasks.

\textit{(1)~Node-level Task.} In a graph $G=(V,E,X)$, given a node $v_i \in V$ with label $y_i$, the node-level task aims to learn a model $\mathcal{F}_{node}(\cdot)$ satisfying $\mathcal{F}_{node}(v_i, G)=y_i$. The node-level dataset is $D_{node} = \{\cdots, ((v_i, G), y_i), \cdots\}$.

\textit{(2)~Edge-level Task.} In a graph $G=(V,E,X)$, given two nodes $v_i, v_j \in V$ with a label $y_{i,j}$, the edge-level task aims to learn a model $\mathcal{F}_{edge}(\cdot)$ satisfying $\mathcal{F}_{edge}(v_i, v_j, G)=y_{i,j}$. The edge-level dataset is $D_{edge} = \{\cdots, ((v_i, v_j, G), y_{i,j}, \cdots\}$.

\textit{(3)~Graph-level Task.} For a graph $G=(V,E,X)$ with a label $y$, the graph-level task aims to learn a model $\mathcal{F}_{graph}(\cdot)$ satisfying $\mathcal{F}_{graph}(G)=y$. The graph-level dataset is $D_{graph} = \{\cdots, ((G_i), y_i), \cdots\}$.

\subsection{Graph Inducing}\label{sec:appendix_graph_ind}

To align three basic graph learning tasks, we follow the existing work~\cite{sun2023all} to reformulate node-level and edge-level tasks as graph-level tasks, respectively, by transforming their inputs into induced graphs, a $\kappa$-ego network. For node-level tasks, an induced graph includes a node and its neighbors within $\kappa$ hops. For edge-level tasks, an induced graph includes two nodes and their neighbors within $\kappa$ hops.

In our experiments, we follow~\cite{sun2023all} construct datasets. For edge-level tasks, we specifically choose samples where the endpoints have identical labels, setting the label of the edge-level sample to match that of its endpoints.
For the graph-level tasks, we select the majority label of the subgraph nodes as the graph-level sample's label. 

\begin{figure}[ht]
    \centering
    \includegraphics[width=\linewidth]{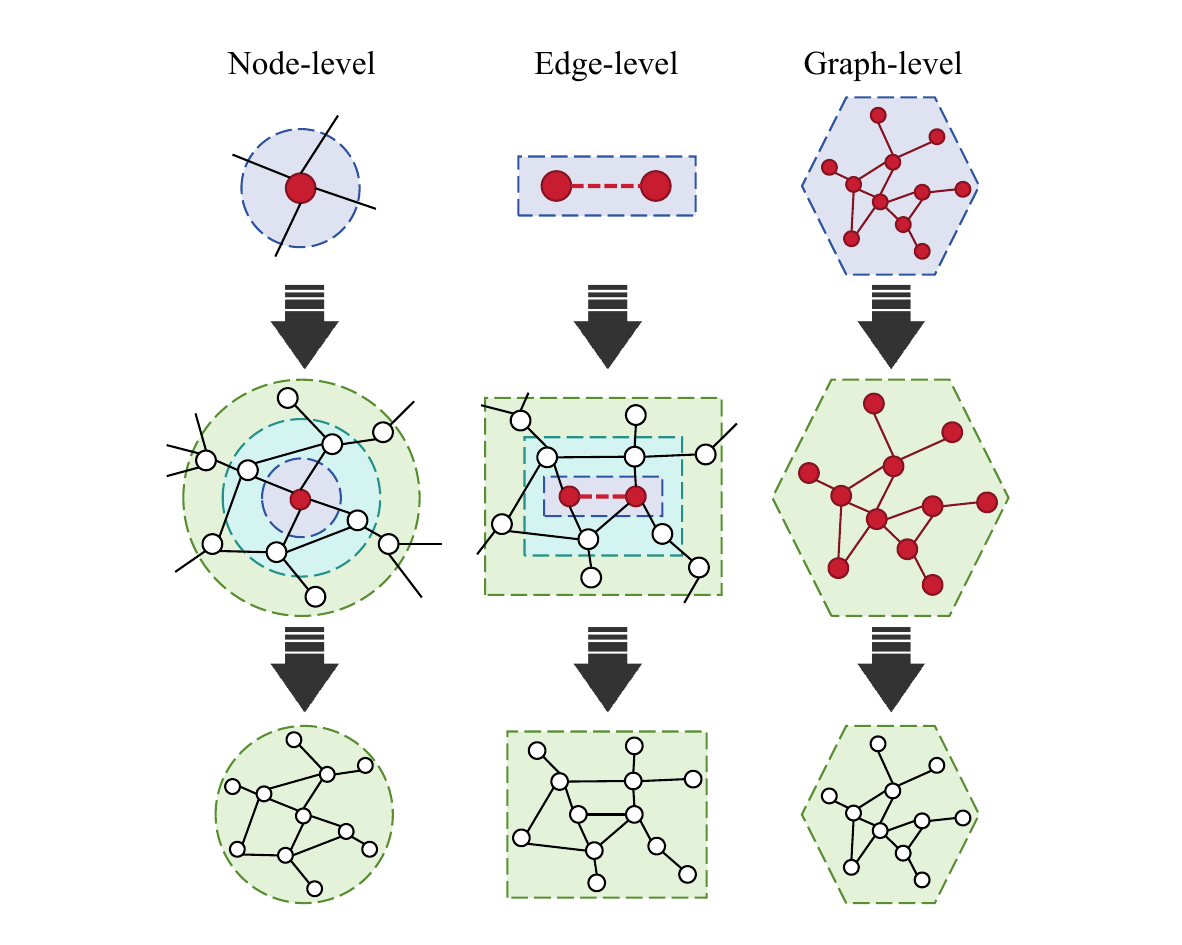}
    \caption{Graph inducing for three levels of tasks.}
    \label{fig:inducing}
\end{figure}

\subsection{Differential Privacy}\label{sec:appendix_dp}

Here we define the Differential Privacy~(DP)~\cite{wei2020federated} for information privatization.
\begin{definition}
    \textbf{Differential Privacy.} A mechanism $\mathcal{M}: \mathcal{D} \rightarrow \mathcal{R}$ that maps domain $\mathcal{M}$ to range $\mathcal{R}$ meets $(\varepsilon ,\delta )$-differential privacy if it satisfies:
    \begin{equation}
        \operatorname{Pr}[\mathcal{M}(d) \in S] \leq e^{\varepsilon} \operatorname{Pr}\left[\mathcal{M}\left(d^{\prime}\right) \in S\right]+\delta
    \end{equation}
    where $d, d^{\prime} \in \mathcal{D}$ are two adjacent inputs, $S\subseteq \mathcal{R}$ are any subset of the mechanism's outputs.
\end{definition}

\subsection{Training Algorithm}\label{sec:appendix_training}

Here we take the node classification task as an example to illustrate the training pipeline in Algorithm~\ref{alg:train}. We set the GNN as a pre-trained and learnable model, which will be optimized in the server after aggregation.

\begin{algorithm}[ht]
  \caption{FedGPL training workflow.}
  \label{alg:train}
  \LinesNumbered
  \KwIn{A server $S$ with a GNN $\mathcal{G}$, and $N$ clients with $f_p(\cdot|\cdot)$'s parameters $\theta_{p}$ and $f_h(\cdot)$'s parameters $\theta_{h}$.}
  \KwOut{Optimized prompt parameters and head parameters for each client.}
  Initialize each client's parameters $\theta_{p}^{(0)}$ and $\theta_{h}^{(0)}$\;
  \For{$n$-th training round not converge}{
    \ForEach{$i$-th client $C_i$ in parallel}{
        $C_i$ prompts all the $k$-th input graphs $\{G_{i,k}\}$ by $\{\tilde{G}_{i,k}\} = \{f_p(G_{i,k}|\hat{G}_i)\}$\;
        $C_i$ sends differentially privatized $\{\tilde{G}_{i,k}\}$ to $S$\;
        $S$ computes representations of $\{\tilde{G}_{i,k}\}$ by $\{h_{i,k}\} = \{\mathcal{G}(\tilde{G}_{i,k})\}$\;
        $S$ sends $\{h_{i,k}\}$ to $C_i$\;
        $C_i$ computes estimated labels $\{\hat{y}_{i,k}\} = \{f_h(h_{i,k})\}$\;
        $C_i$ calculates loss values and enables backward propagation\;
        $C_i$ calculates gradients of $\theta_{p}^{(n-1)}$ and $\theta_{h}^{(n-1)}$\;
        $C_i$ optimizes $\theta_{p}^{(n)}$ and $\theta_{h}^{(n)}$\;
    }
    $S$ calculates $\{\tau_{i \gets j}^{(n+1)}|i,j\in [1,N]\}$ according to $\theta_i^{(n)} = (\theta_{p,i}^{(n)}, \theta_{h,i}^{(n)})$ as Equation~\ref{equ:trans_1} and \ref{equ:trans_2}\;
    $S$ optimizes parameters of $\mathcal{G}$\;
    $S$ sends $\{\tau_{i \gets j}^{(n)}|j\in [1,N]\}$ to $C_i$ for $i \in [1,N]$\;
    \ForEach{$i$-th client $C_i$ in parallel}{
        $C_i$ updates parameters $\theta_{p}^{(n)}$ and $\theta_{h}^{(n)}$ as Equation~\ref{equ:trans_update}\;
    }
  }
\end{algorithm}

\subsection{Proof of Theorem~\ref{the:task_het}}\label{sec:appendix_proof_task_het}

\begin{proof}
    Basically, we suppose that FedGPL consists of two participants as $a$-th and $b$-th clients. Their local parameters are $\theta_a^{(l)}$ and $\theta_b^{(l)}$ at $l$-th step. We can estimate the optimal learning directions by local data as $\theta_a^{(l)\prime}$ and $\theta_b^{(l)\prime}$.
    ACCording to Definition~\ref{def:transferability} we can calculate pairwise transferability as $\tau_{a \gets a}, \tau_{a \gets b}, \tau_{b \gets a}, \tau_{b \gets b}$. Moreover, based on Equations~\ref{equ:trans_1}~and~\ref{equ:trans_2}, we can calculate their parameters after federated aggregation as
    \begin{equation}
        \begin{array}{ll}
            \theta_a^{(l+1)} = & \frac{\tau_{a \gets a}}{\tau_{a \gets a} + \tau_{a \gets b}}\theta_a^{(l)} + \frac{\tau_{a \gets b}}{\tau_{a \gets a} + \tau_{a \gets b}}\theta_b^{(l)}, \\
            \theta_b^{(l+1)} = & \frac{\tau_{b \gets a}}{\tau_{b \gets a} + \tau_{b \gets b}}\theta_a^{(l)} + \frac{\tau_{b \gets b}}{\tau_{b \gets a} + \tau_{b \gets b}}\theta_b^{(l)}. \\
        \end{array}
    \end{equation}
    Next, we assume the learning direction is consistent between two steps, thus we can estimate their optimal learning direction at $(l+1)$-th step as
    \begin{equation}
        \begin{array}{cc}
            \theta_a^{(l+1)\prime} = & \theta_a^{(l+1)} + \overrightarrow{\theta_a^{(l)\prime}-\theta_a^{(l)}}, \\
            \theta_b^{(l+1)\prime} = & \theta_b^{(l+1)} + \overrightarrow{\theta_b^{(l)\prime}-\theta_b^{(l)}}. \\
        \end{array}
    \end{equation}
    The 2-norm parameter difference between two tasks with our proposed aggregation algorithm can be calculated as \begin{equation}
        \begin{array}{ll}
           \Vert \theta_a^{(l+1)\prime} - \theta_b^{(l+1)\prime} \Vert_2 = & \Vert (\frac{\omega_a}{1+\omega_a} + \frac{\omega_b}{1+\omega_b})(\Vert\overrightarrow{\theta_b^{(l)}-\theta_a^{(l)}}\Vert) \\
           & + (\theta_a^{(l+1)} - \theta_b^{(l+1)})\Vert_2,
        \end{array}
    \end{equation}
    where $\omega_a = \frac{\overrightarrow{\theta_a^{(l)\prime}-\theta_a^{(l)}}\cdot\overrightarrow{\theta_b^{(l)\prime}-\theta_a^{(l)}}}{\Vert\overrightarrow{\theta_a^{(l)\prime}-\theta_a^{(l)}}\Vert_2}$ and $\omega_b = \frac{\overrightarrow{\theta_b^{(l)\prime}-\theta_b^{(l)}}\cdot\overrightarrow{\theta_a^{(l)\prime}-\theta_b^{(l)}}}{\Vert\overrightarrow{\theta_b^{(l)\prime}-\theta_b^{(l)}}\Vert_2}$.
    Obviously, if $\Delta_T^{a,b}(\theta_a^{(l+1)\prime},\theta_b^{(l+1)\prime}) 
    \leq \Delta_T^{a,b}(\theta_a^{(l)\prime},\theta_b^{(l)\prime})$ that the task heterogeneity is minor after applying HiDTA, we have
    \begin{equation}
        \begin{array}{rcl}
           \Vert \theta_a^{(l+1)\prime} - \theta_b^{(l+1)\prime} \Vert_2 & \leq &  \Vert \theta_a^{(l)\prime} - \theta_b^{(l)\prime} \Vert_2, \\
           (\frac{\omega_a}{1+\omega_a} + \frac{\omega_b}{1+\omega_b})(\Vert\overrightarrow{\theta_b^{(l)}-\theta_a^{(l)}}\Vert) & \leq & 0, \\
           \frac{\omega_a}{1+\omega_a} + \frac{\omega_b}{1+\omega_b} & \leq & 0,\\
           \frac{1}{\omega_a} + \frac{1}{\omega_b} + 2 & \geq & 0, \\
           \frac{\Vert\overrightarrow{\theta_a^{(l)\prime}-\theta_a^{(l)}}\Vert}{\tau_{a \gets b}} + \frac{\Vert\overrightarrow{\theta_b^{(l)\prime}-\theta_b^{(l)}}\Vert}{\tau_{b \gets a}} + 2 & \geq & 0
        \end{array}
    \end{equation}
    where $\Vert\overrightarrow{\theta_b^{(l)}-\theta_a^{(l)}}\Vert \geq 0$.
    Therefore, when the pairwise transferability values between two clients $\tau_{a \gets b}, \tau_{b \gets a}$ are positive, we have $\Delta_T^{a,b}(\theta_a^{(l+1)\prime},\theta_b^{(l+1)\prime}) 
    \leq \Delta_T^{a,b}(\theta_a^{(l)\prime},\theta_b^{(l)\prime})$, which means that the task heterogeneity decrease after aggregation.
\end{proof}

\subsection{Proof of Theorem~\ref{the:data_het}}\label{sec:appendix_proof_data_het}

\begin{proof}
Given a graph $G(V,E,X)$, we assume that a graph representation computed by a GNN $\mathcal{G}$ follows $h_G \sim U[0,\eta]$. And the GNN is smooth as $h_G = \mathbb{E}_{E \to \emptyset}[\mathcal{G}(V,E,X)]$. The representation of prompted graph $\tilde{G}$ by a VPG $\hat{G}=((\hat{V}^+,\hat{V}^-),(\hat{E}^+,\hat{E}^-),(\hat{X}^+,\hat{X}^-))$ can estimated as
\begin{equation}
    h_{\tilde{G}} = \frac{|V| \cdot h_G + |\hat{V}^+| \cdot h_{G^+} - |\hat{V}^-| \cdot h_G}{|V| + |\hat{V}^+| - |\hat{V}^-|},
\end{equation}
where $h_{G^+} = \mathcal{G}(\hat{V}^+,\hat{E}^+,\hat{X}^+)$.
Next, we assume $\hat{X}^+ \sim X$, $|\hat{V}^+| << |V|$, and $|\hat{V}^+| << |\hat{V}^-|$ in practice, thus we can get the distribution of prompted graphs as
\begin{equation}
    h_{\tilde{G}} \sim \tilde{U}[\eta\alpha_n, \eta],
\end{equation}
where $\alpha_n \in [0,1]$ is a pre-defined percentage parameter of significance score borderline in VPG.

To measure the data heterogeneity between $i$-th and $j$-th clients, we denote their graph data following uniform distributions of $h_{G_a} \sim U^a[0, \eta^a]$ and $h_{G_b} \sim U^b[0, \eta^b]$, respectively. We can infer $h_{\tilde{G}_a} \sim \tilde{U}^a[\eta^a\alpha_n^a, \eta^a]$ and $h_{\tilde{G}_b} \sim \tilde{U}^b[\eta^b\alpha_n^b, \eta^b]$.
Then, the expectation of the 2-norm embedding difference with and without graph prompting $\tilde{\delta}$ can be calculated by
\begin{equation}
    \begin{array}{rl}
         \mathbb{E}[\tilde{\delta}] = & \mathbb{E}_{\left\{\substack{h_{\tilde{G}_a} \sim \tilde{U}^a\\h_{\tilde{G}_b} \sim \tilde{U}^b}\right.}
    [\Vert h_{\tilde{G}_a} - h_{\tilde{G}_b} \Vert_2] - \mathbb{E}_{\left\{\substack{h_{G_a} \sim U^a\\h_{G_b} \sim U^b}\right.}[\Vert h_{G_a} - h_{G_b} \Vert_2]\\
        = & ((\frac{1-\alpha_n^a}{2}\eta^a)^2+(\frac{1-\alpha_n^b}{2}\eta^b)^2-2(\frac{1-\alpha_n^a}{2}\eta^a)(\frac{1-\alpha_n^b}{2}\eta^b))\\
        & - ((\frac{1}{2}\eta^a)^2+(\frac{1}{2}\eta^b)^2-2(\frac{1}{2}\eta^a)(\frac{1}{2}\eta^b))\\
        = & \frac{(1-\alpha_n^a)^2-1}{4}{\eta^a}^2+\frac{(1-\alpha_n^b)^2-1}{4}{\eta^b}^2-2(\frac{(1-\alpha_n^a)(1-\alpha_n^b)-1}{4}\eta^a\eta^b).
    \end{array}
\end{equation}
Obviously, when $\alpha_n^a \approx \alpha_n^b$, we ensure
\begin{equation}
    \mathbb{E}[\tilde{\delta}] \approx \frac{(1-\alpha_n^a)^2-1}{4}(\eta_a-\eta_b)^2 \leq 0,
\end{equation}
because $(1-\alpha_n^a)^2-1 \leq 0$.
Subsequently, according to Equation~\ref{equ:task_het} that is monotonically increasing, the data heterogeneity can be reduced after graph prompting as
\begin{equation}
    \mathbb{E}_{\left\{\substack{h_{\tilde{G}_a} \sim \tilde{U}^a\\h_{\tilde{G}_b} \sim \tilde{U}^b}\right.}[\Delta_{D}^{i,j}(h_{\tilde{G}_a},h_{\tilde{G}_b})] \leq \mathbb{E}_{\left\{\substack{h_{G_a} \sim U^a\\h_{G_b} \sim U^b}\right.}[\Delta_{D}^{i,j}(h_{G_a},h_{G_b})].
\end{equation}
\end{proof}

\subsection{Dataset Statistics}\label{sec:appendix_dataset}

Table~\ref{tab:datasets} are the statistics of graph datasets, including Cora, CiteSeer, DBLP, Photo, and Physics, which are transformed from traditional graph learning datasets for node classification.

\begin{table}[ht]
    \centering
    \caption{Statistics of datasets.}
    \begin{tabular}{@{}c|cccc@{}}
    \toprule
    Dataset  & \# Nodes & \# Edges & \# Features & \# Labels \\ \midrule
    Cora     & $2,708$    & $5,429$    & $1,433$       & $7$        \\
    CiteSeer & $3,327$    & $9,104$    & $3,703$       & $6$        \\
    DBLP     & $17,716$   & $105,734$  & $602$        & $6$        \\
    Photo    & $7,650$    & $238,162$  & $745$        & $8$        \\
    Physics  & $34,493$   & $495,924$  & $8,415$       & $5$        \\ \bottomrule
    \end{tabular}
    \label{tab:datasets}
\end{table}

\subsection{Implementation Details}\label{sec:appendix_imp}
For the implementation of federated learning, we deploy $9$ clients, and each level of tasks (node-level, edge-level, and graph-level) contains $3$ clients, each client has $400$ induced graphs on average, which are induced from the raw dataset with $\kappa=5$, and each client participates in every communication round, during each round, clients undergo a single epoch of training, and the global communication round is set to $50$. 
For the implementation of graph prompting methods, for GPF, we add additional feature vectors into the node features, for ProG, we set the number of tokens as $10$. Each client contains its training dataset, prompt, and an answering layer that projects the embedding output by the GNN to the final results. 
The learning rate is set as $0.1$ for all datasets. The results are averaged on all clients. The HiFGL framework is implemented based on PyTorch~\cite{paszke2019pytorch}, and PyTorch-Lightning~\cite{falcon2019pytorch} runs on the machine with Intel Xeon Gold 6148 @ 2.40GHz, V100 GPU and 64G memory.

\begin{table*}[ht]
\setlength{\tabcolsep}{4pt}
\caption{Overall performance~(ACC~($\%$) and F1~($\%$)) of different federated algorithms and graph prompting methods in few-shot settings.}
\begin{tabular}{@{}c|c|
>{\columncolor[HTML]{EFEFEF}}c 
>{\columncolor[HTML]{EFEFEF}}c |cc|
>{\columncolor[HTML]{EFEFEF}}c 
>{\columncolor[HTML]{EFEFEF}}c |cc|
>{\columncolor[HTML]{EFEFEF}}c 
>{\columncolor[HTML]{EFEFEF}}c @{}}
\toprule
                                                                             &                                                                           & \multicolumn{2}{c|}{\cellcolor[HTML]{EFEFEF}Cora} & \multicolumn{2}{c|}{CiteSeer}   & \multicolumn{2}{c|}{\cellcolor[HTML]{EFEFEF}DBLP} & \multicolumn{2}{c|}{Photo}      & \multicolumn{2}{c}{\cellcolor[HTML]{EFEFEF}Physics} \\
\multirow{-2}{*}{\begin{tabular}[c]{@{}c@{}}Federated\\ Method\end{tabular}} & \multirow{-2}{*}{\begin{tabular}[c]{@{}c@{}}Prompt\\ Method\end{tabular}} & ACC                     & F1                      & ACC            & F1             & ACC                     & F1                      & ACC            & F1             & ACC                      & F1                       \\ \midrule
                                                                             & GPF                                                                       & $80.06$                   & $79.84$                   & $82.34$          & $81.84$          & $78.46$                   & $78.15$                   & $74.51$          & $73.14$          & $86.15$                    & $86.21$                    \\
                                                                             & ProG                                                                      & $84.25$                   & $84.31$                   & $83.19$          & $83.04$          & $79.84$                   & $79.73$                   & $69.84$          & $69.47$          & $85.89$                    & $85.03$                    \\
\multirow{-3}{*}{Local}                                                      & VPG                                                                    & $85.81$                   & $85.10$                   & $84.52$          & $84.37$          & $82.31$                   & $82.17$                   & $77.97$          & $76.84$          & $88.03$                    & $88.14$                    \\ \midrule
                                                                             & GPF                                                                       & $79.81$                   & $79.57$                   & $82.79$          & $82.87$          & $79.64$                   & $78.84$                   & $79.03$          & $78.14$          & $86.87$                    & $86.65$                    \\
                                                                             & ProG                                                                      & $80.44$                   & $80.26$                   & $84.54$          & $83.81$          & $80.14$                   & $80.20$                   & $70.12$          & $70.03$          & $86.84$                    & $86.67$                    \\
\multirow{-3}{*}{FedAvg}                                                     & VPG                                                                    & $86.14$                   & $85.82$                   & $85.01$          & $84.77$          & $80.72$                   & $80.07$                   & $79.31$          & $79.01$          & $88.72$                    & $88.12$                    \\ \midrule
                                                                             & GPF                                                                       & $77.37$                   & $76.41$                   & $81.04$          & $80.97$          & $78.07$                   & $77.91$                   & $80.57$          & $80.19$          & $84.58$                    & $84.29$                    \\
                                                                             & ProG                                                                      & $78.59$                   & $78.14$                   & $80.08$          & $80.14$          & $77.48$                   & $77.97$                   & $70.16$          & $70.07$          & $85.17$                    & $85.01$                    \\
\multirow{-3}{*}{FedProx}                                                    & VPG                                                                    & $84.32$                   & $84.23$                   & $81.24$          & $80.88$          & $78.14$                   & $78.45$                   & $84.25$          & $84.12$          & $90.05$                    & $90.03$                    \\ \midrule
                                                                             & GPF                                                                       & $72.94$                   & $72.44$                   & $72.48$          & $70.84$          & $77.61$                   & $77.31$                   & $72.15$          & $70.13$          & $82.14$                    & $82.23$                    \\
                                                                             & ProG                                                                      & $70.67$                   & $68.47$                   & $60.41$          & $60.87$          & $75.14$                   & $75.10$                   & $68.72$          & $68.12$          & $83.17$                    & $83.04$                    \\
\multirow{-3}{*}{SCAFFOLD}                                                   & VPG                                                                    & $79.74$                   & $79.55$                   & $75.85$          & $84.76$          & $77.22$                   & $77.07$                   & $73.17$          & $72.94$          & $86.14$                    & $86.07$                    \\ \midrule
                                                                             & GPF                                                                       & $80.68$                   & $80.81$                   & $82.99$          & $82.83$          & $80.75$                   & $80.66$                   & $84.10$          & $84.07$          & $87.02$                    & $87.06$                    \\
\multirow{-2}{*}{HiDTA}                                                      & ProG                                                                      & $80.22$                   & $79.28$                   & $83.48$          & $83.45$          & $80.67$                   & $80.17$                   & $69.47$          & $69.38$          & $88.87$                    & $88.17$                    \\
\midrule
\multicolumn{2}{c|}{FedGPL}                                                                                       & \textbf{86.45}          & \textbf{86.42}          & \textbf{85.59} & \textbf{85.56} & \textbf{82.72}          & \textbf{81.87}          & \textbf{85.28} & \textbf{85.26} & \textbf{90.14}           & \textbf{90.09}\\
\bottomrule
\end{tabular}
\label{tab:overall_fewshot}
\end{table*}

\subsection{Overall Performance for Supervised and Prompt Learning}

We evaluate the performance by training modules from scratch~(\ie Supervised) and freezing the pre-trained GNN~(\ie Prompt) in Table~\ref{tab:supervised_prompt}. Beyond fine-tuning, the results of the two training schemes depict that FedGPL behaves more competitively than others on two datasets.

\begin{table}[ht]
\caption{ACC~($\%$) and F1~($\%$) of different federated algorithms and graph prompting methods. Supervised: training models from scratch; Prompt: fine-tuning models based on a frozen pre-trained GraphTransformer.}
\label{tab:supervised_prompt}
\begin{tabular}{@{}c|cc|cc|cc@{}}
\toprule
\multirow{2}{*}{\begin{tabular}[c]{@{}c@{}}Training\\ Scheme\end{tabular}} & \multicolumn{1}{c|}{\multirow{2}{*}{\begin{tabular}[c]{@{}c@{}}Federated\\ Method\end{tabular}}} & \multirow{2}{*}{\begin{tabular}[c]{@{}c@{}}Prompt\\ Method\end{tabular}} & \multicolumn{2}{c|}{Cora}       & \multicolumn{2}{c}{CiteSeer}    \\
                                                                           & \multicolumn{1}{c|}{}                                                                            &                                                                          & ACC            & F1             & ACC            & F1             \\ \midrule
\multirow{7}{*}{Training}                                                  & \multicolumn{1}{c|}{\multirow{3}{*}{FedAvg}}                                                     & GPF                                                                      & 85.67          & 85.34          & 84.62          & 84.2           \\
                                                                           & \multicolumn{1}{c|}{}                                                                            & ProG                                                                     & 87.96          & 87.61          & 85.71          & 85.87          \\
                                                                           & \multicolumn{1}{c|}{}                                                                            & SUPT                                                                     & 86.31          & 86.24          & 85.98          & 85.51          \\ \cmidrule(l){2-7} 
                                                                           & \multicolumn{1}{c|}{\multirow{3}{*}{FedProx}}                                                    & GPF                                                                      & 81.34          & 80.51          & 83.22          & 82.71          \\
                                                                           & \multicolumn{1}{c|}{}                                                                            & ProG                                                                     & 85.61          & 85.40          & 84.41          & 83.85          \\
                                                                           & \multicolumn{1}{c|}{}                                                                            & SUPT                                                                     & 82.31          & 82.45          & 83.49          & 83.8           \\ \cmidrule(l){2-7} 
                                                                           & \multicolumn{2}{c|}{FedGPL}                                                                                                                                                 & \textbf{89.14} & \textbf{88.57} & \textbf{88.12} & \textbf{87.94} \\ \midrule
\multirow{7}{*}{Prompt}                                                    & \multicolumn{1}{c|}{\multirow{3}{*}{FedAvg}}                                                     & GPF                                                                      & 80.50          & 79.72          & 83.22          & 82.71          \\
                                                                           & \multicolumn{1}{c|}{}                                                                            & ProG                                                                     & 84.02          & 83.77          & 84.41          & 83.85          \\
                                                                           & \multicolumn{1}{c|}{}                                                                            & SUPT                                                                     & 81.83          & 80.61          & 83.49          & 83.8           \\ \cmidrule(l){2-7} 
                                                                           & \multicolumn{1}{c|}{\multirow{3}{*}{FedProx}}                                                    & GPF                                                                      & 77.81          & 76.97          & 82.61          & 82.33          \\
                                                                           & \multicolumn{1}{c|}{}                                                                            & ProG                                                                     & 82.54          & 81.81          & 82.64          & 81.67          \\
                                                                           & \multicolumn{1}{c|}{}                                                                            & SUPT                                                                     & 78.67          & 78.34          & 82.45          & 82.37          \\ \cmidrule(l){2-7} 
                                                                           & \multicolumn{2}{c|}{FedGPL}                                                                                                                                                 & \textbf{87.61} & \textbf{87.13} & \textbf{87.58} & \textbf{87.74} \\ \bottomrule
\end{tabular}
\end{table}

\subsection{Overall Performance with Pre-trained GCN}

We test the fine-tuned FGL performance based on a pre-trained GCN on Cora and Citeseer datasets in Table~\ref{tab:gcn}. FedGPL outperform baseline models in terms of ACC and F1.

\begin{table}[ht]
\caption{ACC~($\%$) and F1~($\%$) of different federated algorithms and graph prompting methods based on a pre-trained GCN~\cite{kipf2016semi}.}
\label{tab:gcn}
\begin{tabular}{@{}cc|cc|cc@{}}
\toprule
\multicolumn{1}{c|}{\multirow{2}{*}{\begin{tabular}[c]{@{}c@{}}Federated\\ Method\end{tabular}}} & \multirow{2}{*}{\begin{tabular}[c]{@{}c@{}}Prompt\\ Method\end{tabular}} & \multicolumn{2}{c|}{Cora}       & \multicolumn{2}{c}{CiteSeer}    \\
\multicolumn{1}{c|}{}                                                                            &                                                                          & ACC            & F1             & ACC            & F1             \\ \midrule
\multicolumn{1}{c|}{\multirow{3}{*}{Local}}                                                      & GPF                                                                      & 83.14          & 82.64          & 81.63          & 81.37          \\
\multicolumn{1}{c|}{}                                                                            & ProG                                                                     & 84.91          & 84.33          & 82.21          & 81.44          \\
\multicolumn{1}{c|}{}                                                                            & SUPT                                                                     & 82.41          & 82.01          & 82.14          & 82.07          \\ \midrule
\multicolumn{1}{c|}{\multirow{3}{*}{FedAvg}}                                                     & GPF                                                                      & 84.61          & 84.37          & 84.24          & 83.67          \\
\multicolumn{1}{c|}{}                                                                            & ProG                                                                     & 85.61          & 85.03          & 84.81          & 84.02          \\
\multicolumn{1}{c|}{}                                                                            & SUPT                                                                     & 83.17          & 83.34          & 83.61          & 83.44          \\ \midrule
\multicolumn{1}{c|}{\multirow{3}{*}{FedProx}}                                                    & GPF                                                                      & 79.47          & 79.31          & 82.67          & 82.33          \\
\multicolumn{1}{c|}{}                                                                            & ProG                                                                     & 83.44          & 82.97          & 83.41          & 83.01          \\
\multicolumn{1}{c|}{}                                                                            & SUPT                                                                     & 78.67          & 78.34          & 82.76          & 82.03          \\ \midrule
\multicolumn{1}{c|}{\multirow{3}{*}{SCAFFOLD}}                                                   & GPF                                                                      & 76.61          & 75.51          & 72.51          & 72.03          \\
\multicolumn{1}{c|}{}                                                                            & ProG                                                                     & 72.61          & 72.81          & 73.67          & 73.11          \\
\multicolumn{1}{c|}{}                                                                            & SUPT                                                                     & 73.35          & 72.85          & 72.61          & 72.17          \\ \midrule
\multicolumn{2}{c|}{FedGPL}                                                                                                                                                 & \textbf{86.87} & \textbf{86.27} & \textbf{85.34} & \textbf{85.22} \\ \bottomrule
\end{tabular}
\end{table}

\begin{table*}[ht]
\centering
\caption{Comparison of ACC~($\%$) for different graph prompting methods and training schemes. N, E, and G:~node-, edge-, and graph-level task, (f):~few-shot settings.}
\begin{tabular}{@{}c|ccc|ccc|ccc@{}}
\toprule
\rowcolor[HTML]{FFFFFF} 
\cellcolor[HTML]{FFFFFF}                         & \multicolumn{3}{c|}{\cellcolor[HTML]{FFFFFF}Cora} & \multicolumn{3}{c|}{\cellcolor[HTML]{FFFFFF}CiteSeer} & \multicolumn{3}{c}{\cellcolor[HTML]{FFFFFF}Physics} \\
\rowcolor[HTML]{FFFFFF} 
\multirow{-2}{*}{\cellcolor[HTML]{FFFFFF}Method} & N               & E              & G              & N                & E                & G               & N               & E               & G               \\ \midrule
\rowcolor[HTML]{FFFFFF} 
Supervised                                       & $80.72$           & $95.21$          & $95.14$          & $81.00$            & $97.45$            & $88.83$           & $83.07$           & $92.78$           & $99.78$           \\
\rowcolor[HTML]{FFFFFF} 
Pre-train                                        & $77.06$           & $87.52$          & $88.61$          & $80.68$            & $93.19$            & $72.25$           & $75.44$           & $78.94$           & $95.36$           \\
\rowcolor[HTML]{FFFFFF} 
Fine-tune                                        & $81.01$           & $95.36$          & $95.57$          & $81.33$            & $97.27$            & $85.67$           & $83.81$           & $92.47$           & $99.87$           \\ \midrule
\rowcolor[HTML]{FFFFFF} 
GPF                                              & $76.17$           & $88.24$          & $87.05$          & $82.11$            & $93.81$            & $74.17$           & $76.51$           & $86.66$           & $98.42$           \\
\rowcolor[HTML]{FFFFFF} 
ProG                                             & $77.56$           & $89.49$          & $89.48$          & $82.51$            & $95.57$            & $75.94$           & $76.42$           & $87.12$           & $97.45$           \\
\rowcolor[HTML]{FFFFFF} 
VPG                                              & {$79.19$}     & {$90.35$}    & {$91.02$}    & {$82.94$}      & {$96.48$}      & {$77.49$}     & {$83.37$}     & {$91.03$}     & {$99.50$}     \\ \midrule
\rowcolor[HTML]{EFEFEF} 
GPF(f)                                           & $74.35$           & $82.33$          & $85.62$          & $81.02$            & $92.88$            & $73.11$           & $75.27$           & $79.74$           & $96.04$           \\
\rowcolor[HTML]{EFEFEF} 
ProG(f)                                          & $75.14$           & $87.51$          & $89.57$          & $81.41$            & $93.43$            & $74.75$           & $80.78$           & $78.41$           & $97.47$           \\
\rowcolor[HTML]{EFEFEF} 
VPG(f)                                           & {$77.64$}     & {$89.46$}    & {$90.35$}    & {$81.89$}      & {$94.82$}      & {$76.84$}     & {$82.14$}     & {$83.78$}     & {$98.17$}     \\ \bottomrule
\end{tabular}
\label{tab:prompt}
\end{table*}

\subsection{Privacy Analysis}

\begin{figure}[ht]
    \centering
    \includegraphics[width=0.7\linewidth]{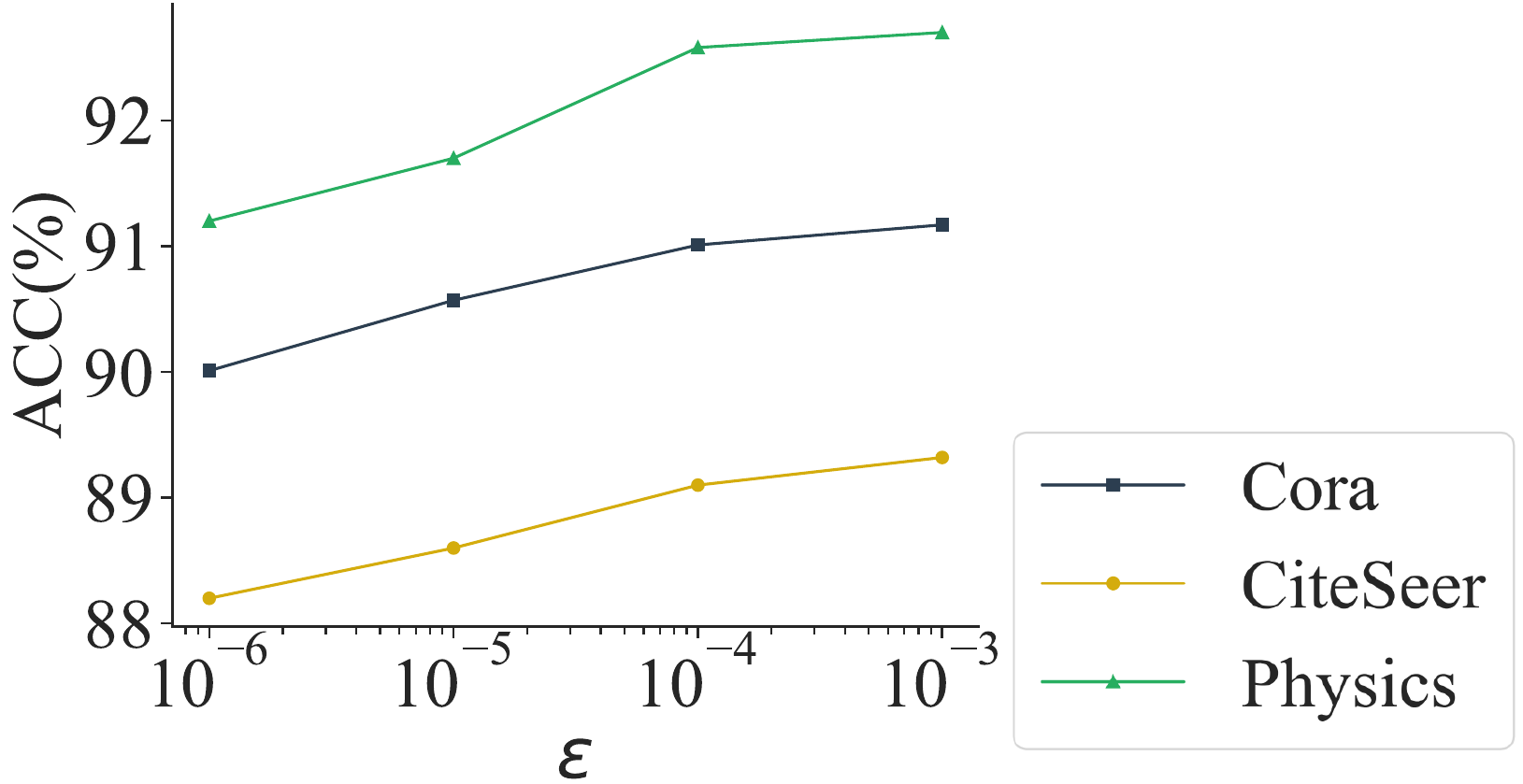}
    \caption{Performance-privacy correlation curves.}
    \label{fig:privacy}
\end{figure}

We evaluate the impact of using differential privacy on the performance of FedGPL. Here we conduct experiments by varying the privacy scale of $\epsilon$ and test it on the Cora dataset.
The results are shown in Figure~\ref{fig:privacy}, which demonstrates the relationship between ACC and privatization parameters. We observe that the ACC is reduced as we increase the privacy constraints (decreasing $\epsilon$). Similar conclusions are drawn from different settings. The finding indicates that there is a trade-off between performance and privacy in FedGPL, which is common in FL~\cite{zhang2023trading}. In practical utilization, we will choose proper $\epsilon$ for different degrees of privacy protection demands on GNN and graph data.

\subsection{Case Study: Transferability}
\begin{figure}[ht]
    \centering
    \includegraphics[width=0.5\linewidth]{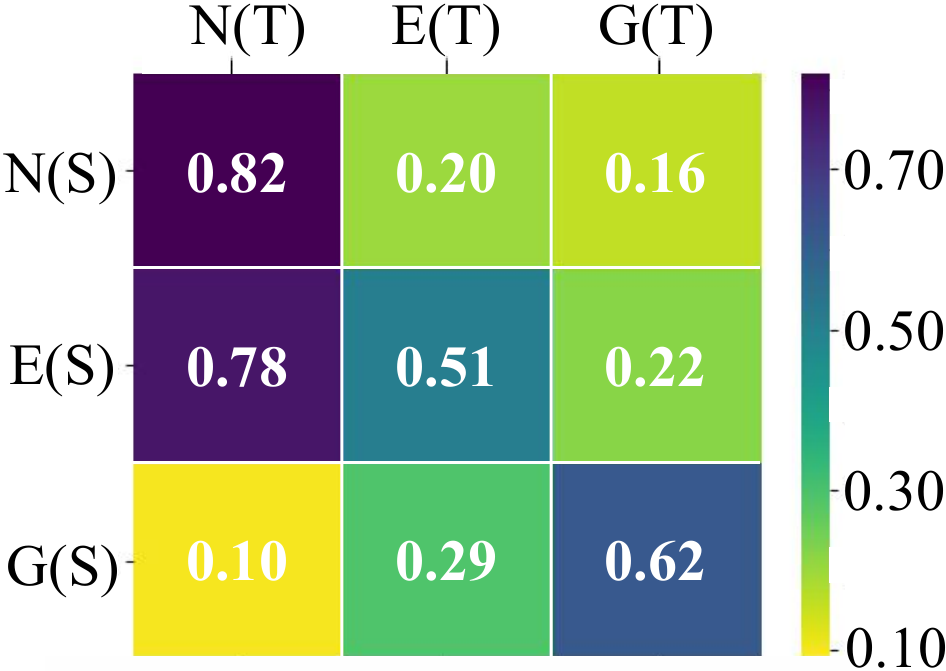}
    \caption{Transferability. N, E, and G:~node-, edge-, and graph-level task, S:~source task, and T:~target task.}
    \label{fig:heatmap2}
\end{figure}

To further discuss how HiDTA becomes robust against two kinds of heterogeneity, we conduct a case study on transferability according to the aggregating weight matrix, to find out how transferability models the heterogeneous knowledge transfer. Specifically, Figure~\ref{fig:heatmap2} shows the averaged and normalized transferability values in the last $20$ training epochs on CiteSeer. The results suggest that HiDTA prioritizes learning directions with higher transferability. Besides, the optimization process of a target task mainly depends on itself, aligned with our motivations that aim to model asymmetric knowledge transfer.

\subsection{Performance on Few-shot Settings}

Besides the standard setting with full training samples, we follow GPL works~\cite{sun2023all} to deploy a few-shot setting~\textbf{(f)}. Few-shot settings are usually incorporated in prompt learning, which acts as an efficient learning manner. It lets each client have only a limited number of data samples.
In our experiments, we utilize a $100$-shot scenario, where each client possesses only $100$ labeled samples. The prediction results are shown in Table~\ref{tab:overall_fewshot}.
We observe that FedGPl still outperforms baselines in few-shot settings in a prompt tuning way, which demonstrates that its generalization ability successfully address FGL in a task- and data- heterogeneous scenario.

\subsection{Experiments on Graph Prompting}

To further evaluate the effectiveness of VPG, the graph prompting module of FedGPL, we compare VPG with GPF and ProG under the local graph prompting setting on Cora, CiteSeer, and Physics datasets, as shown in Table~\ref{tab:prompt}. We also compared with (1)~supervised training: training a graph model from scratch, (2)~fine-tuning: fine-tuning the GNN with labeled data, and (3)~pre-training: freezing GNN with only a learnable task head.
We observe that VPG performs better than ProG and GPF by $0.43\%$ to $7.28\%$ across all tasks on three datasets. Generally, prompting unleashes more potential of GNN on different downstream tasks with small-scale trainable parameters. In some cases, prompting even surpasses fine-tuning and supervised learning, which may be attributed to a lightweight model for more effective optimization.

\begin{table*}[ht]
\centering
\caption{Comparison of ACC~($\%$) for different graph prompting methods and federated algorithms. N, E, and G:~node-, edge-, and graph-level task.}
\setlength{\tabcolsep}{5pt}
\begin{tabular}{@{}cc|ccc|ccc|ccc@{}}
\toprule
\multicolumn{1}{c|}{\multirow{2}{*}{\begin{tabular}[c]{@{}c@{}}Federated\\ Method\end{tabular}}} & \multirow{2}{*}{\begin{tabular}[c]{@{}c@{}}Prompt\\ Method\end{tabular}} & \multicolumn{3}{c|}{Cora}                        & \multicolumn{3}{c|}{CiteSeer}                    & \multicolumn{3}{c}{Physics}                      \\ \cmidrule(l){3-11} 
\multicolumn{1}{c|}{}                                                                            &                                                                          & N              & E              & G              & N              & E              & G              & N              & E              & G              \\ \midrule
\multicolumn{1}{c|}{\multirow{3}{*}{Local}}                                                      & GPF                                                                      & $76.17$          & $88.24$          & $87.05$          & $82.11$          & $93.81$          & $74.17$          & $76.51$          & $86.66$          & $98.42$          \\
\multicolumn{1}{c|}{}                                                                            & ProG                                                                     & $77.56$          & $89.49$          & $89.48$          & $80.74$          & $91.57$          & $73.03$          & $78.09$          & $80.83$          & $98.03$          \\
\multicolumn{1}{c|}{}                                                                            & VPG                                                                      & $79.19$          & $90.35$          & $91.05$          & $82.94$          & $96.48$          & $77.49$          & $83.37$          & $91.03$          & $99.51$          \\ \midrule
\multicolumn{1}{c|}{\multirow{3}{*}{FedAvg}}                                                     & GPF                                                                      & $75.87$          & $87.21$          & $86.74$          & $82.44$          & $93.71$          & $75.03$          & $77.94$          & $88.96$          & $98.57$          \\
\multicolumn{1}{c|}{}                                                                            & ProG                                                                     & $77.61$          & $89.14$          & $90.13$          & $82.86$          & $86.42$          & $86.27$          & $79.74$          & $82.84$          & $99.13$          \\
\multicolumn{1}{c|}{}                                                                            & VPG                                                                      & $79.34$          & $90.57$          & $91.26$          & $83.31$          & $96.78$          & $78.21$          & $83.42$          & $91.21$          & $99.34$          \\ \midrule
\multicolumn{1}{c|}{\multirow{3}{*}{FedProx}}                                                    & GPF                                                                      & $72.21$          & $82.84$          & $83.11$          & $82.86$          & $87.33$          & $75.86$          & $75.64$          & $87.24$          & $96.48$          \\
\multicolumn{1}{c|}{}                                                                            & ProG                                                                     & $75.84$          & $87.31$          & $85.74$          & $81.03$          & $91.91$          & $73.13$          & $81.87$          & $83.13$          & $98.32$          \\
\multicolumn{1}{c|}{}                                                                            & VPG                                                                      & $76.61$          & $88.16$          & $86.81$          & $81.37$          & $94.47$          & $76.61$          & $83.25$          & $91.12$          & $99.04$          \\ \midrule
\multicolumn{1}{c|}{\multirow{3}{*}{SCAFFOLD}}                                                   & GPF                                                                      & $69.28$          & $78.48$          & $79.51$          & $75.41$          & $79.18$          & $69.37$          & $72.57$          & $84.47$          & $94.42$          \\
\multicolumn{1}{c|}{}                                                                            & ProG                                                                     & $58.97$          & $62.41$          & $65.48$          & $61.47$          & $68.81$          & $64.23$          & $73.51$          & $85.24$          & $94.57$          \\
\multicolumn{1}{c|}{}                                                                            & VPG                                                                      & $70.57$          & $77.04$          & $75.52$          & $72.61$          & $84.76$          & $67.11$          & $79.88$          & $87.34$          & $95.21$          \\ \midrule
\multicolumn{1}{c|}{\multirow{2}{*}{HiDTA}}                                                      & GPF                                                                      & $74.13$          & $85.91$          & $86.01$          & $81.47$          & $94.51$          & $76.68$          & $79.92$          & $87.87$          & $98.82$          \\
\multicolumn{1}{c|}{}                                                                            & ProG                                                                     & $74.24$          & $86.04$          & $86.11$          & $82.83$          & $95.84$          & $77.74$          & $82.55$          & $88.16$          & $98.72$          \\ \midrule
\multicolumn{2}{c|}{FedGPL}                                                                                                                                                 & \textbf{82.19} & \textbf{92.49} & \textbf{92.02} & \textbf{83.41} & \textbf{96.81} & \textbf{78.41} & \textbf{83.86} & \textbf{91.79} & \textbf{99.47} \\ \bottomrule
\end{tabular}
\label{tab:detailed}
\end{table*}

\subsection{Hyperparameter Sensitivity}
\begin{figure}[ht]
    \centering
    \includegraphics[width=\linewidth]{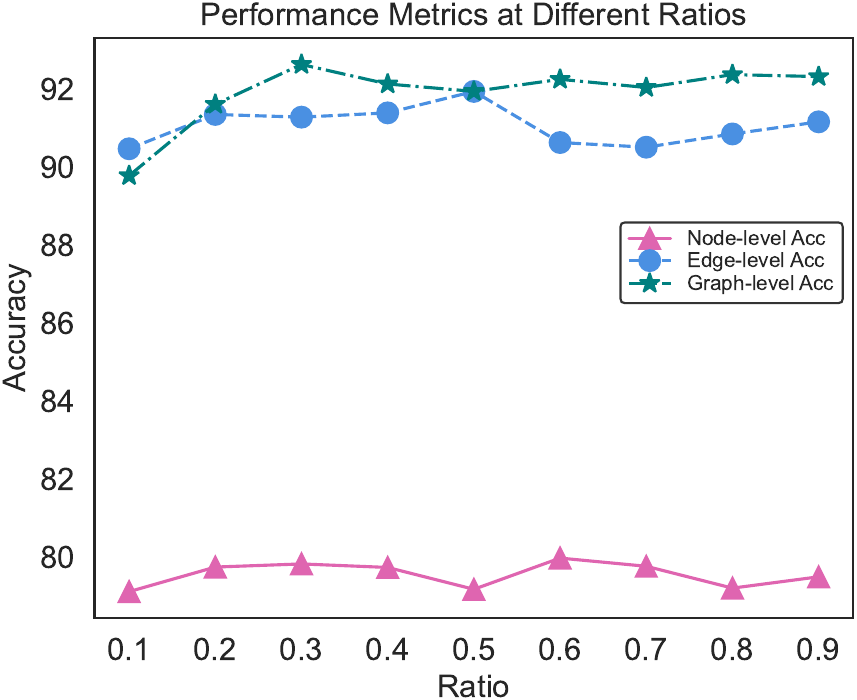}
    \caption{Parameter sensitivity of $\alpha_n$.}
    \label{fig:hyper}
\end{figure}

We also investigate the impact of different $\alpha_n$, as shown in Figure~\ref{fig:hyper}, where we conclude that optimal $\alpha_n$ of tasks are distinct.
For example, $\alpha_n=0.3$ is optimal for the graph-level task, while $\alpha_n=0.5$ is optimal for the edge-level task. Results further reveal the existence of heterogeneity between different levels of tasks. This heterogeneity indicates personalized prompting is essential and motivates us to design a selective cross-client knowledge sharing method to improve prediction accuracy.

\subsection{Task Performance}

Table~\ref{tab:detailed} is a detailed comparison of the accuracy of different federated learning algorithms combined with graph prompting methods on three datasets across three types of tasks. From the table, it is evident that FedGPL outperforms other methods in all three tasks. For example, on the Cora dataset, compared to local training, FedGPL shows improvements of $4.02$\%, $2.14$\%, and $0.97$\% in node-level, edge-level, and graph-level tasks, respectively. This demonstrates that FedGPL enhances prompting performance by facilitating knowledge sharing across different tasks. On the other hand, compared to FedAvg, HiDTA exhibits better performance with VPG. For instance, on the Cora dataset, HiDTA shows improvements of $3.00$\%, $2.14$\%, and $0.97$\% in the three tasks compared to FedAvg, indicating that HiDTA better mitigates the heterogeneity between different tasks, enabling more effective knowledge sharing.

\end{document}